\documentclass{article}

\usepackage{microtype}
\usepackage{graphicx}
\usepackage{subcaption}
\usepackage{booktabs} % for professional tables

\usepackage{hyperref}

\usepackage[accepted]{stylefile}

\usepackage{amsmath}
\usepackage{amssymb}
\usepackage{mathtools}
\usepackage{amsthm}

\usepackage[capitalize,noabbrev]{cleveref}

\theoremstyle{plain}

\theoremstyle{definition}

\theoremstyle{remark}

\usepackage[textsize=tiny]{todonotes}

\usepackage{dsfont}
\usepackage{placeins}

\usepackage{balance}

\newcommand{\beginsupplement}{ % use to mark beginning of supplementary section. 
        \setcounter{section}{0}
        \setcounter{table}{0}
        \renewcommand{\thetable}{S\arabic{table}} %
        \setcounter{figure}{0}
        \renewcommand{\thefigure}{S\arabic{figure}} %
     }

\renewcommand\cite{\citep} % make citep the default behavior

\newcommand{\probclassifier}[1]{\widehat{p}_{\mathcal{M}}(Y_{#1} \mid X_{#1} )}

\newcommand{\probclassifierklower}[1]{\widehat{p}_{\mathcal{M}}(Y_{#1} = k \mid X = x_{#1} )}
\newcommand{\probclassifierensemble}[2]{\widehat{p}_{\mathcal{M}_{#2}}(Y_{#1} \mid X_{#1} )}
\newcommand{\probclassifiermeanshort}{\widehat{p}_{\bar{\mathcal{M}},i,k}}
\newcommand{\probclassifiershort}{\widehat{p}_{\mathcal{M},i,k}}
\newcommand{\probclassifiershortscaled}{\widehat{p}_{\mathcal{M},i,k}^{(T)}}
\newcommand{\probclassifiershortk}[1]{\widehat{p}_{\mathcal{M},i,{#1}}}
\newcommand{\probclassifierensembleshort}[1]{\widehat{p}_{\mathcal{M}_{#1},i,k}}
\newcommand{\probempk}[1]{\widehat{p}_\text{emp}(Y_{#1} = k \mid \{ Y_{ij} \}_{j \in \mathcal{J}_i} )}

\newcommand{\annotatorpostshortk}[2]{\widehat{p}_{\mathcal{A}_j,{#1},{#2}}}
\newcommand{\annotatorpostsim}{\widehat{p}_{\mathcal{A}_j}}

\DeclareMathOperator*{\argmax}{arg\,max}

\newcommand{\gitlink}{\url{https://github.com/cleanlab/multiannotator-benchmarks/tree/main/active_learning_benchmarks}}

\newcommand{\papertitle}{ActiveLab: Active Learning with Re-Labeling by Multiple Annotators}

\newcommand{\appendixtitle}{Appendix \\[0.6em] \papertitle}

\icmltitlerunning{\papertitle}

\begin{document}

\twocolumn[
\icmltitle{\papertitle}


\icmlsetsymbol{equal}{*}

\begin{icmlauthorlist}
\icmlauthor{Hui Wen Goh}{cleanlab}
\icmlauthor{Jonas Mueller}{cleanlab}
\end{icmlauthorlist}

\icmlaffiliation{cleanlab}{Cleanlab}

\icmlcorrespondingauthor{Hui Wen Goh}{huiwen@cleanlab.ai}
\icmlcorrespondingauthor{Jonas Mueller}{jonas@cleanlab.ai}

\icmlkeywords{Machine Learning, ICML, Active Learning}

\vskip 0.3in
]



\printAffiliationsAndNotice{\icmlEqualContribution} % otherwise use the standard text.

\begin{abstract}
In real-world data labeling applications, annotators often provide imperfect labels. It is thus common to employ multiple annotators to label data with some overlap between their examples. We study active learning in such settings, aiming to train an accurate classifier by collecting a dataset with the fewest total  annotations. Here we propose ActiveLab, a practical method to decide what to label next that works with any classifier model and can be used in pool-based batch active learning with one or multiple annotators. ActiveLab automatically estimates when it is more informative to re-label examples vs.\  labeling entirely new ones. This is a key aspect of producing high quality labels and trained models within a limited annotation budget.  In experiments on image and tabular data, ActiveLab reliably trains more accurate classifiers with far fewer annotations than a wide variety of popular active learning methods. 

\end{abstract}

\section{Introduction}

Model-agnostic active learning methods  use outputs from some \emph{arbitrary} type of trained prediction model in order to identify the \emph{most informative} data to label, so that a more accurate version of the same model can be trained. 
Such general approaches are popular because they can be directly applied to many data modalities (image, text, etc.) as long as a reasonable model can be trained. 
Focusing on highly practical settings, we consider model-agnostic pool-based active learning with multiple data annotators that label a batch of many examples in between model (re)training runs. This setting is easy to setup and allows us to address common issues in real-world active learning such as: labelers who are imperfect, or expensive model (re)training that cannot be executed every time a new example is labeled.
% We believe this to be the most practically-relevant setting for active learning in practice, where one can typically not guarantee individual annotators will provide perfectly accurate labels and model training may be too expensive to do one new example at a time.
Working with annotators that may provide incorrect labels, it is useful to sometimes ask new annotators to provide extra labels for examples previously labeled by others. This allows us to verify the current consensus label or estimate a better one. % \looseness=-1

Here we introduce ActiveLab\footnote{Our code: \gitlink{}}, a straightforward \textbf{active} learning algorithm that estimates \emph{when} such re-\textbf{lab}eling will be more effective than labeling an entirely new example. A very general approach, ActiveLab can be used:  with any type of classifier model (or ensemble of multiple models) and data modality, for active learning with multiple annotators where the set of annotators changes over time, for traditional active learning where each example is labeled at most once (Appendix  \ref{sec:singlelabel}), and for active label cleaning where all data is already labeled by at least one annotator and the goal is to establish the highest quality consensus labels within a limited annotation budget. % by having additional annotators re-label certain difficult/suspicious examples.
ActiveLab is easy-to-implement and computationally efficient.

% The novel approach of both labeling new examples and re-labeling a previously labeled examples aims to not only obtain more information from labeling additional examples, but also correcting any label errors from previously labeled examples through obtaining a more reliable consensus label through multiple annotators (ie. having a larger number of annotators label a more difficult example would increase the likelihood of that example getting an accurate label).

\section{Methods}

This paper focuses on classification tasks with $K$ classes, 
for which some (arbitrary) classifier model $\mathcal{M}$ can be trained. For our $i$th example with feature values $X_i$, this model predicts a class probability vector $\probclassifier{i}$ estimating the likelihood that $X$ belongs to each class $k \in [K] := \{1,2,\dots, K\}$.

In the \emph{pool}-based \emph{batch} active learning settings we consider,  each round involves the steps described below. 
% Some notation is formally defined later on.
In the beginning, we start with a training set $\mathcal{D}$ of examples that have at least one (noisy) annotation, where some of these examples may have been labeled by multiple annotators. We also have a pool of unlabeled examples $\mathcal{U}$ that have zero annotations. Our proposed active learning method may choose to collect new labels for examples in either $\mathcal{D}$ or $\mathcal{U}$. Based on classifier predictions $\widehat{p}$ and the currently-observed annotations $\mathcal{D}$, ActiveLab estimates an acquisition score $s_i$ for each example. Examples with the lowest $s_i$ values are those for which collecting an additional label is expected to be most informative when subsequently training $\mathcal{M}$.
To avoid overfit/biased results, classifier predictions  $\widehat{p}$ should be \emph{out-of-sample}, coming from a copy of the model $\mathcal{M}$ that has never been trained with the example it is asked to predict the class of.

\renewcommand{\algorithmicrequire}{\textbf{Input:}}
\begin{algorithm}
\caption*{Active learning with multiple annotators}\label{alg:euclid}
\begin{algorithmic}[1]
\REQUIRE $\mathcal{D}$: labeled examples with at least one annotation 
\REQUIRE $\mathcal{U}$: unlabeled pool of examples (not yet annotated)
\FOR{$r = 1,2, \dots$ \COMMENT{rounds of active learning}}
\STATE Estimate consensus labels $\widehat{Y}_i$ for annotated examples $x_i \in \mathcal{D}$ (some of which have multiple annotations) \label{step:consensus}
\STATE Train classifier model $\mathcal{M}_r$ with these labels: $(x_i, \widehat{Y}_i)$\looseness=-1 \label{step:train}
\STATE Obtain (out-of-sample) predicted class probabilities for all examples: $\widehat{p} = \mathcal{M}_r(x)$ for $x \in \mathcal{D} \cup \mathcal{U}$ \label{step:predict}
\STATE Use active learning method to score all examples: $ s_i = A(\widehat{p}_i;\mathcal{D})$ for all $x_i \in \mathcal{D} \cup \mathcal{U}$ \label{step:score}
\STATE Assemble batch $\mathcal{B}$ of the $B$ best-scoring examples, collect \textbf{one} additional label $Y_{ij}$ for each  $x_i \in \mathcal{B}$, and add new $\{Y_{ij}\}$ to the training data (updating $\mathcal{D}, \mathcal{U}$) \label{step:collect}
\ENDFOR % \label{endforloop}
\end{algorithmic}
\end{algorithm}
% \vspace*{-3mm}

We can obtain out-of-sample predictions for every  $x_i \in \mathcal{D}$ by fitting our model via $k$-fold cross-validation in Step \ref{step:train}. For examples currently in the unlabeled pool $x \in \mathcal{U}$, 
 Step \ref{step:collect} can collect their first label, and there may be already-labeled examples $x \in \mathcal{D}$ in the selected batch  $\mathcal{B}$ for which we collect yet another label. 
There are many ways to operationalize the collection of labels in Step \ref{step:collect} of active learning. The examples to acquire an extra label for could be divided amongst a limited pool of annotators (some of which labeled other examples in previous active learning rounds), or these examples could be given to new annotators to label.

While one can envision alternate methods that suggest which annotator should label which example \cite{huang2017cost}, we find such a tightly-controlled setting too rigid for many applications. Step \ref{step:collect} is intentionally flexible. 
We also do not consider methods that can ask more than one annotator to review the same example
within a round as such methods can be brittle  % (vs.\ our methods which collect at most one new label for each example in a round)
\cite{baldridge2004active}. 
% Beyond the setting outlined here, our ActiveLab algorithm can also be applied in settings where: there is no unlabeled pool $\mathcal{U}$, also known as \emph{active label cleaning} \cite{bernhardt2022active}, or only new labels are collected for examples from $\mathcal{U}$ rather than sometimes re-labeling examples from $\mathcal{D}$ (i.e.\ traditional active learning), or only a single annotator is available to label the data.

% Each round of active learning involves the following steps: (1) Obtain consensus labels for the examples that have at least one annotation (only the training set at the beginning); (2)  Train a classifier model with these consensus labels; (3) Evaluate the test accuracy of this classifier against ground truth labels (only used for evaluation purposes); (4) Use this classifier to obtain out-of-sample predicted class probabilities for all examples in train and unlabeled pools; (5) Apply an active learning method to estimate the acquisition value for each example; (6) Collect one additional label for each example that belongs to the batch with the $B$ best acquisition values (this may be the first label for examples from $\mathcal{U}$), and add these new labels to the training data before repeating step (1). 
\vspace*{-1em}
\paragraph{Notation.} % We now define the remaining notation, here omitting subscripts $r$ for objects that change in each round and defining them with respect to the current round. 
In the remaining notation, all definitions of objects are  given with respect to the current round. Here we omit subscripts $r$ and how objects change between rounds. 
In the current round, the set of annotated examples $\mathcal{D}$ contains $n$ examples labeled by $m$ annotators in total. $Y_{ij} \in [K]$ denotes the class annotator $\mathcal{A}_j$ chose for example $x_i \in \mathcal{D}$, with $Y_{ij} = \emptyset$ if annotator $\mathcal{A}_j$ did not label example $i$. $\mathcal{Y}_i$ is the set of collected labels for example $x_i$, with $| \mathcal{Y}_i | = 0$ if $x_i \in \mathcal{U}$. 
$\mathcal{I}_j$ is the subset of examples labeled by annotator $\mathcal{A}_j$, and $\mathcal{J}_i$ is the subset of annotators that labeled $x_i$. 
% Our methods estimate an acquisition score $s_i$, inversely related to the value of collecting an additional label for  example $x_i$. In each round of active learning, we  prioritize labeling the examples with the lowest scores.

\subsection{ActiveLab}
\label{sec:activelab}

ActiveLab extends the CROWDLAB estimator of \citet{crowdlab}. Some equations in this paper overlap with CROWDLAB, but we present them for completeness. Not every CROWDLAB equation is motivated here, curious readers can refer to detailed explanations by \citet{crowdlab}. 

Unlike ActiveLab, which is intended for guiding collection of additional labels, CROWDLAB is intended for analyzing a static dataset labeled by multiple annotators. Empirically it performs poorly when used for active learning. While both approaches estimate consensus labels in a similar fashion, they score  examples differently. CROWDLAB estimates the likelihood that each current consensus label is \emph{correct} or not, whereas ActiveLab estimates the utility of collecting \emph{another} label to further improve the consensus and model trained therewith. CROWDLAB assigns very low scores to examples annotated by many labelers that heavily disagree, but even though their consensus label is unreliable, ActiveLab recognizes there is less utility in collecting one more label for such fundamentally difficult examples (vs.\ examples that currently have fewer annotations). 
Unlike CROWDLAB, ActiveLab also scores examples which currently have not been labeled yet. It must trade-off the potential information gain from collecting the 1st label for an example from $\mathcal{U}$ vs.\ the $j$th label for an example already labeled $j-1$  times. Both methods can utilize any type of classifier model $\mathcal{M}$ trained in any fashion.

We first describe how ActiveLab computes the score $s_i$ for examples that have at least one annotation. 
Both CROWDLAB and ActiveLab are straightforward weighted ensembles which linearly combine multiple predictors to form a single estimate of class probabilities. In prediction competitions, such ensembles are often more accurate and better calibrated. One of these predictors is the (out-of-sample predictions from a) trained classifier $\mathcal{M}$, abbreviated as $\probclassifiershortk{k} := \probclassifierklower{i}$. The other predictors are the annotators who previously labeled $x_i$. From the label $Y_{ij}$ chosen by annotator $\mathcal{A}_j$, we form an   annotator-estimated class probability vector $\annotatorpostshortk{i}{k} \approx p(Y_i = k \mid Y_{ij})$ that is directly comparable to the classifier predicted class probabilities (details further below). ActiveLab and CROWDLAB take a weighted average of this collection of probabilistic predictions to form a single vector of ensemble predicted class probabilities for each $x_i$. 

CROWDLAB subsequently selects the most likely class under this ensemble estimate as the consensus label $\widehat{Y}_i$ representing our best guess of the true label $Y_i$. 
In Step \ref{step:consensus} of each active learning round, we use CROWDLAB to estimate a single \emph{consensus label} $\widehat{Y}_i$ that aggregates the available annotations $\mathcal{Y}_i$ for each example $x_i \in \mathcal{D}$. 
Subsequently in Step \ref{step:score}, ActiveLab scores $x_i \in \mathcal{D}$ via the likelihood that class $\widehat{Y}_i$ is correct under its ensemble estimate, expressed as: \looseness=-1
\begin{align}
    & \text{If } x_i \in \mathcal{D}: \label{eq:al_d} \\ \nonumber
    & s_i =  % \nonumber \\
    \frac{w_{\mathcal{M}} \cdot \probclassifiershortk{\widehat{Y}_i} + w_{\bar{\mathcal{A}}} \cdot \frac{1}{K} + \sum_{j \in \mathcal{J}_i} w_{j} \cdot  \annotatorpostshortk{i}{\widehat{Y}_i} }{
    w_{\mathcal{M}} + w_{\bar{\mathcal{A}}} + \sum_{j \in \mathcal{J}_i} w_{j}}
    \\
    & \text{If } x_i \in \mathcal{U}: \ \ 
    s_i = \frac{w_{\mathcal{M}} \cdot \max_k \probclassifiershort{} + w_{\bar{\mathcal{A}}} \cdot \frac{1}{K}}{w_{\mathcal{M}} + w_{\bar{\mathcal{A}}}}  \label{eq:al_u}
\end{align}
The above estimates depend on $w_\mathcal{M}, w_{j}$ which  determine \emph{how much} to weigh the model $\mathcal{M}$ and each annotator $\mathcal{A}_j$. We estimate their relative trustworthiness (based on the observed annotations $\{Y_{ij}\}$) in order to select these weights, via the same procedure as CROWDLAB (details further below). Intuitively our estimate should  down-weigh untrustworthy annotators or a poorly trained classifier, see \citet{crowdlab} for further discussion on this estimate's robustness against bad  annotators/models. 
Unlike CROWDLAB, equation (\ref{eq:al_d}) also contains a  uniform $1/K$ predictor that  receives weight $w_{\bar{\mathcal{A}}} := \frac{1}{m} \sum_{j=1}^m  w_{j}$, representing the weight assigned to our average annotator (across all examples). 

Here is a fundamental difference between ActiveLab and CROWDLAB: under the former, the estimated likelihood that $\widehat{Y}_i$ is the correct class for $x_i \in \mathcal{D}$ is much lower (closer to uniform) for examples with few annotations. This regularization has smaller effect on examples with many annotations. Thus amongst the $x \in \mathcal{D}$, ActiveLab naturally favors acquiring labels for examples that currently have fewer annotations. ActiveLab also favors examples where annotators disagree with the consensus (note $\annotatorpostshortk{i}{\widehat{Y}_i}$ is much smaller if $Y_{ij} \neq \widehat{Y}_i$) or the classifier predicts the consensus to be unlikely. These are the $x_i \in \mathcal{D}$ whose current consensus label may be wrong, warranting re-labeling to determine whether a better label can be established. 

\paragraph{Scoring examples from the unlabeled pool.} Before delving into the details of $w_\mathcal{M}, w_{j}$, and $\widehat{p}_{\mathcal{A}_j}$, we describe how ActiveLab scores $x_i \in \mathcal{U}$. This is detailed in equation (\ref{eq:al_u}). Since we have no annotations for $x_i \in \mathcal{U}$, ActiveLab scores such examples only using the probabilistic predictions from our classifier $\widehat{p}_\mathcal{M}$. Many traditional active learning methods also operate this way \citep{munro2021human}. As seen in (\ref{eq:al_u}), the score $s_i$ for $x_i \in \mathcal{U}$ is similarly computed as for $x_i \in \mathcal{D}$, except for  modifications required to handle missing information. Since $\mathcal{J}_i = \emptyset$ in this case, we simply drop the annotator-predictors $\widehat{p}_{\mathcal{A}_j}$ from the weighted ensemble in order to obtain its estimate for unlabeled examples. And we simply take $\widehat{Y}_i = \argmax_k \probclassifiershort{}$, the class predicted by our classifier, since CROWDLAB cannot estimate a consensus label for $x_i \in \mathcal{U}$. Amongst the unlabeled examples, ActiveLab thus favors acquiring labels for those $x_i$ for which the classifier is least confident, as in traditional  uncertainty sampling \cite{munro2021human}.

\paragraph{To label or re-label?} Since they are computed in a similar fashion, the $s_i$ are directly comparable between $x_i \in \mathcal{D}$ vs.\ $ \mathcal{U}$. ActiveLab thus naturally suggests when it is better to \textbf{re-label} an example from $\mathcal{D}$ vs.\ labeling a \textbf{new} example from $\mathcal{U}$.
Cases when this might be true for some example $x_i \in \mathcal{D}$ include settings where: its annotations disagree (indicating that some annotators are noisy), or the model has atypically low confidence in its prediction (indicating $x_i$ may be an outlier or high-influence datapoint whose label we should really get right), or the model confidently disagrees with the annotations. This last case is especially pertinent for examples $x_i$ that only have a single annotation, where we may prefer to trust a confident prediction from a well-trained classifier over the given label which may be wrong \cite{northcutt2021labelerrors,labelerrordetection}.
Fixing labels for existing training data can improve a classifier more than noisily labeling additional data  \cite{northcutt2021confidentlearning, iraola2021single}. Section \ref{sec:single_vs_multi} empirically explores this.  
Mathematically, it is evident that ActiveLab will always prefer to label new examples from $\mathcal{U}$ if every annotation and the classifier (confidently) agree for all $x_i \in \mathcal{D}$.

\paragraph{Example.} Consider $x_i$ with a single annotation $Y_{ij}$ and a different $x_{\ell} \in \mathcal{U}$, such that our classifier is equally confident in its predictions for both. In this case, deciding whether to re-label $x_i$ vs.\ labeling $x_{\ell}$ specifically depends on: whether $Y_{ij}$ matches the classifier's predicted class $\argmax_k \probclassifiershort{}$, and how much ActiveLab weights this annotator ($w_j$) vs.\ the average annotator ($w_{\bar{\mathcal{A}}}$) and the classifier ($w_{\mathcal{M}}$). If the classifier's prediction matches $Y_{ij}$, then ActiveLab will prefer to label $x_{\ell}$.
If the classifier disagrees with the annotation, then ActiveLab will prefer to re-label $x_{i}$ whenever the CROWDLAB consensus label $\widehat{Y}_i \neq Y_{ij}$. This occurs if:
\begin{equation*}
w_{\mathcal{M}} \big(\probclassifiershortk{k^*} - \probclassifiershortk{Y_{ij}} \big) > w_j \big( \annotatorpostshortk{i}{Y_{ij}} - \annotatorpostshortk{i}{k^*} \big)
% \label{eq:examplerelabel}
\end{equation*}
where $k^* := \argmax_k \probclassifiershort{} \neq Y_{ij}$ in this example. 
The inequality is satisfied if: 
$w_{\mathcal{M}} \gg w_j$ (i.e.\ ActiveLab estimates the classifier is more trustworthy than annotator $\mathcal{A}_j$) and $\probclassifiershortk{k^*} - \probclassifiershortk{Y_{ij}} \gg \annotatorpostshortk{i}{Y_{ij}} - \annotatorpostshortk{i}{k^*}$ (i.e.\ the classifier predicts $Y_{ij}$ is not the correct label confidently relative to the estimated accuracy of the data annotators).

\paragraph{Details for estimating weights and annotator likelihood.}
ActiveLab estimates $w_\mathcal{M}, w_{j}$, and $\annotatorpostsim$ in the same fashion as CROWDLAB. We present the mathematical details here but refer readers to the explanations/motivations articulated by \citet{crowdlab}. 
In equation (\ref{eq:al_d}),  $\annotatorpostsim \in \mathbb{R}^k$ is an ``annotator likelihood'' vector  containing the probabilities that $x_i$ belongs to each class given that annotator $\mathcal{A}_j$ chose the label $Y_{ij}$.
It is very simply defined:
\begin{align*}
    \annotatorpostshortk{i}{k} \approx& \ \ p(Y_i = k \mid Y_{ij})  
    :=& \begin{cases}
    P & \mbox{when } Y_{ij} = k\\
    \frac{1 - P}{K - 1} & \mbox{when } Y_{ij} \neq k
\end{cases}
\end{align*}
 $P \ge 0$ is a global scalar parameter shared across all annotators. It is estimated by computing the average annotator agreement across all examples that have more than one annotation. $P$ estimates the probability that a typical annotator would select the consensus label for an arbitrary example they are given \cite{crowdlab}. % \looseness=-1
\iffalse
\begin{gather*}
    P = \frac{1}{|\mathcal{I}_+|} \sum_{i \in \mathcal{I}_+} \frac{1}{| \mathcal{J}_i |} \sum_{j \in \mathcal{J}_i} \mathds{1} ( Y_{ij} = \widehat{Y}_i ) \\
    \text{ where } \mathcal{I}_+ := \{i \in [n] : |\mathcal{J}_i| > 1 \}
\end{gather*}
\fi

The weights $w_\mathcal{M}, w_{j}$ in equation (\ref{eq:al_d}) estimate the trustworthiness of our classifier model and each annotator. 
The model weight is defined in terms of the normalized accuracy of the classifier's predictions with respect to the consensus label, over the subset of examples with more than one annotation. The weight $w_j$ for annotator $\mathcal{A}_j$ is defined in terms of how much labels chosen by $\mathcal{A}_j$ agree with other annotators when they labeled the same examples as $\mathcal{A}_j$. More formally: \looseness=-1
\begin{align*}
    w_{j} & := 1 - \frac{1 - g_j}{1 - A_{\text{MLC}}} \\
    w_{\mathcal{M}} & := \left( 1 - \frac{1 - A_{\mathcal{M}}}{1 - A_{\text{MLC}}} \right) \cdot \sqrt{\frac{1}{n} \sum_{i \in \mathcal{D}} |\mathcal{J}_i|}
\end{align*}
Above $g_j$ is the a\textbf{g}reement between  $\mathcal{A}_j$ and other annotators:
\begin{equation*}
    g_j := \frac{\sum_{i \in \mathcal{I}_j} \sum_{\ell \in \mathcal{J}_i, \ell \neq j} \mathds{1} (Y_{ij} = Y_{i\ell})}{\sum_{i \in \mathcal{I}_j} ( |\mathcal{J}_i| -1 )}
\end{equation*}
$A_{\mathcal{M}}$ represents the empirical accuracy of the classifier model's predictions with respect to the consensus labels:
\begin{equation}
    A_{\mathcal{M}} := \frac{1}{|\mathcal{I}_+|} \sum_{i \in \mathcal{I}_+} \mathds{1} \bigg(\widehat{Y}_i = \argmax_k \probclassifiershortk{k} \bigg) 
    \label{eq:am}
\end{equation}
$A_{\text{MLC}}$ is a normalization factor, the baseline accuracy (with respect to consensus labels) achieved by  
 predicting the overall most labeled class $Y_{\text{MLC}}$ (amongst all annotations for the dataset) always for every example. 
\begin{equation}
    A_{\text{MLC}} := \frac{1}{|\mathcal{I}_+|} \sum_{i \in \mathcal{I}_+} \mathds{1} (Y_{\text{MLC}} = \widehat{Y_i})
    \label{eq:amlc}
\end{equation}
Note that to avoid bias \cite{crowdlab}, the accuracy estimates which determine $P$, $w_{j}$, and $w_{\mathcal{M}}$ are always computed over the subset of labeled examples that received more than one annotation:
$\mathcal{I}_+ := \{i \in \mathcal{D} : |\mathcal{J}_i| > 1 \} $.

\subsection{Calibration of Classifier Predictions}
\label{sec:calibration}

While cross-validation enables us to produce out-of-sample predictions for each $x_i \in \mathcal{D}$, some types of models tend to nonetheless output overconfident predictions \cite{guo2017calibration}. 
Our active learning methods rely on the classifier to determine what data to label next and subsequently retrain another version of this same classifier. In this self-reinforcing process, overconfident predictions may be extremely detrimental. 
% A classifier model's overconfidence of its predictions may be extremely detrimental to the active learning process as the ActiveLab algorithm relies on the classifier model's predicted probabilities from its previous active learning iteration to determine each example's \emph{active learning quality score}.

To mitigate overconfidence (or underconfidence), we   \emph{calibrate} the classifier's predicted class probabilities in Step \ref{step:predict} of each active learning round, before we compute ActiveLab scores via  equation (\ref{eq:al_d}). We perform this calibration against the empirical distribution of the annotators'  labels $\mathcal{Y}_i$ for each example in $\mathcal{D}$. 
% This empirical distribution serves as a crude measure of the confidence in each example's true label. 
Calibration is done by temperature scaling \cite{guo2017calibration} the classifier's predicted probabilities $\probclassifier{i}$ to minimize their (soft) cross entropy against the empirical distribution $\widehat{p}_{\text{emp}}$ of classes in $\mathcal{Y}_i$. That is, we choose the temperature $T$ to maximize: 
\begin{align*} 
    & \sum_{i \in \mathcal{D}} \ \sum_{k=1}^K \probempk{i} \cdot \log \probclassifiershortscaled{} \\
   \text{ where } & \ \probclassifiershortscaled{} = \sigma \hspace*{-0.5mm} \left(  \hspace*{-0.5mm} \frac{\probclassifiershort}{T} \hspace*{-0.5mm} \right) \text{for softmax } \sigma(z_k) = \frac{e^{z_k}}{\sum_k e^{z_k}}
\end{align*}
After identifying the best value of $T$, we calibrate the predictions for all examples in both $\mathcal{D}$ and $\mathcal{U}$ and compute ActiveLab scores using  $\probclassifiershortscaled{}$ in place of $\probclassifiershort$. 
In our experiments, this calibration step improved a variety of active learning methods, allowing them to more robustly improve the accuracy of various types of models.

\subsection{ActiveLab (Ensemble)}

Ensemble methods aggregate outputs from multiple (independently trained) models
into a single set of predictions that can be more accurate than any of the constituent models \cite{dietterich2000ensemble}.
Model ensembles are also popular in active learning; disagreeing predictions between models indicate areas of high epistemic uncertainty where annotating more data can greatly improve at least one of the constituent models \cite{seung1992query}. Here we present an straightforward extension of ActiveLab to ensemble settings. 

Assuming there are $L$ trained models in an ensemble, let $\probclassifierensemble{i}{\ell}$ denote the class probabilities for $x_i$ predicted by model $\mathcal{M}_\ell$ for $\ell = 1,2,...,L$. Here we can apply ActiveLab similarly as in the single-model case, but now allowing each model to have its own weight $w_{\mathcal{M}_1}$, $w_{\mathcal{M}_2}, ..., w_{\mathcal{M}_L}$ used for averaging estimates. 
We use the following ActiveLab scores in ensemble settings:
\begin{align}
    & \text{If } x_i \in \mathcal{D}: \label{eq:alens_d} \\ \nonumber
    & s_i =  % \nonumber \\
    \frac{ w_{\bar{\mathcal{A}}} \cdot \frac{1}{K} + 
 \sum\limits_{\ell=1}^L w_{\mathcal{M}_\ell} \cdot \widehat{p}_{\mathcal{M}_\ell, i, \widehat{Y}_i}  + \sum\limits_{j \in \mathcal{J}_i} w_{j} \cdot  \annotatorpostshortk{i}{\widehat{Y}_i} }{ w_{\bar{\mathcal{A}}} + 
    \sum\limits_{\ell=1}^L w_{\mathcal{M}_\ell} + \sum\limits_{j \in \mathcal{J}_i} w_{j}}
    \\[0.5em]
    & \text{If } x_i \in \mathcal{U}: \ \ 
    s_i = \frac{ w_{\bar{\mathcal{A}}} \cdot \frac{1}{K} + \sum_{\ell=1}^L w_{\mathcal{M}_\ell} \cdot \widehat{p}_{\mathcal{M}_\ell, i, \widetilde{Y}_i} }{ w_{\bar{\mathcal{A}}} + \sum_{\ell=1}^L w_{\mathcal{M}_\ell} }  \label{eq:alens_u}
\end{align}
Above the annotator weights $w_j, w_{\bar{\mathcal{A}}}$ and likelihoods $\widehat{p}_{\mathcal{A}_j}$ have the same definitions as in ActiveLab with a single model. 
Here consensus labels $\widehat{Y}_i$ are estimated from an similar ensemble extension of CROWDLAB, in which we propose to  set $w_{\bar{\mathcal{A}}} = 0$ in equation (\ref{eq:alens_d}) and identify which class $\widehat{Y}_i \in [K]$ maximizes the expression. 
Equation (\ref{eq:alens_u}) shows we handle examples from the unlabeled pool in the same fashion as in the single-model case. For each $x_i \in \mathcal{U}$, we obtain a predicted class $\widetilde{Y}_i \in [K]$ from the ensemble classifier and treat $\widetilde{Y}_i$ as a proxy for its consensus label.

% describing model weights
The weights $w_{\mathcal{M}_\ell}$ for each model are  computed the same way as in ActiveLab with a single model. Each model's prediction accuracy with respect to consensus labels is again used to infer how trustworthy each model is relative to the annotators, with $A_{\mathcal{M}_\ell}$ and $A_{\text{MLC}}$ defined as in (\ref{eq:am}) and (\ref{eq:amlc}).
\begin{equation*}
    w_{\mathcal{M}_\ell} := \left( 1 - \frac{1 - A_{\mathcal{M}_\ell}}{1 - A_{\text{MLC}}} \right) \cdot \sqrt{\frac{1}{n} \sum_i |\mathcal{J}_i|}
\end{equation*}
To predict with our ensemble classifier after training, we can also take a weighted average of each model's predicted class probabilities using the same weights $w_{\mathcal{M}_\ell}$. \looseness=-1

\iffalse
% OLD
we can get a proxy for the consensus label by taking the majority vote label based on the predictions of the multiple classifier models, this proxy consensus label will be denoted as $\widetilde{Y}_i$.

single-model active learning quality score above, the score for the examples that have at least one annotator can be computed by

\begin{gather}
    s_i = \nonumber \\
     \frac{\sum_{r=1}^R w_{\mathcal{M}_r} \cdot \probclassifiershortk{\widehat{Y}_i} + \sum_{j \in \mathcal{J}_i} w_{j} \cdot  \annotatorpostshortk{i, \widehat{Y}_i} + w_{\bar{\mathcal{A}}} \cdot \frac{1}{K}}{\sum_{r=1}^R w_{\mathcal{M}_r} + \sum_{j \in \mathcal{J}_i} w_{j} + w_{\bar{\mathcal{A}}}}  
\end{gather}

For the examples without any annotator labels, we can get a proxy for the consensus label by taking the majority vote label based on the predictions of the multiple classifier models, this proxy consensus label will be denoted as $\widetilde{Y}_i$.

Similar to the single-model case, these examples will not have any associated annotator likelihood probability vector. Hence the quality score is computed by

\begin{equation}
    s_i = \frac{\sum_{r=1}^R w_{\mathcal{M}_r} \cdot \probclassifiershortk{\widetilde{Y}_i} + w_{\bar{\mathcal{A}}} \cdot \frac{1}{K}}{\sum_{r=1}^R w_{\mathcal{M}_r} + w_{\bar{\mathcal{A}}}}  
\end{equation}
\fi

\section{Related Work}

The most popular active learning methods are those like ActiveLab that can used with any classifier model for any data modality  \cite{munro2021human}. While there has been extensive research on active learning \cite{zhan2022comparative} and analyzing crowdsourced labels \cite{paun2018comparing}, few model/modality-agnostic active learning methods have been developed for settings with multiple annotators and data re-labeling \cite{lin2014re}. 
Many of the active learning methods proposed for such settings are specific to certain types of models or data types  \cite{rodrigues2014gaussian,zhao2011incremental,yan2011active, yang2018leveraging, huang2017cost, gilyazev2018active, iraola2021single}. Other approaches like impact sampling \cite{lin2016re} are too computationally expensive to run on problems like the image classification task in Section \ref{sec:experiments}.

\subsection{Baseline Methods}

Our subsequent experiments benchmark ActiveLab against the following commonly used  model/modality-agnostic methods for active learning and data re-labeling. Each method is applied in the same manner as ActiveLab to iteratively label a dataset, except which $x_i$ are  labeled is chosen via different $s_i$, and consensus labels for all $x_i \in \mathcal{D}$ are computed via majority-vote as used by \citet{zheng2010active}.

\textbf{Random}.  
This method selects which examples to annotate entirely at random. It uses score: $s_i = x$ where $x \in [0, 1]$ is sampled uniformly at random and independently of $i$. 

\textbf{Good Random}.  
This is a better variant of random selection that accounts for the number of annotations $x_i$ already has: $s_i = x + | \mathcal{Y}_i |$ 
where $x \in [0, 1]$ is sampled uniformly at random. This pseudo-random selection prioritizes examples with the fewest number of labels collected thus far, a simpler variant of the approach of \citet{chen2022clean}. The unlabeled pool is labeled first prior to any re-labeling.

\textbf{Entropy} \cite{cohn1996active}.  
This method scores examples via the entropy of the model-predicted probabilities.
\begin{equation}
    s_i = \sum_{k=1}^K  \probclassifiershort{} \cdot \log \probclassifiershort{}
\end{equation}

\textbf{Uncertainty} \cite{cohn1996active}.  
Measures how confident the model is in its predicted class: \ $s_i = \max_k \probclassifiershort{}$.

\textbf{Active Label Cleaning} \cite{bernhardt2022active}.  
This approach was recently proposed for efficiently re-labeling an already-labeled dataset with multiple annotators. To select which data to collect an extra annotation for, \citet{bernhardt2022active} introduce a  score that is a difference of two terms. The first term is the cross-entropy between the $\mathcal{M}$-predicted class probabilities and the empirical distribution of the annotators' labels for a particular example, and the second term is the entropy of the $\mathcal{M}$-predicted class  probabilities. 
\begin{align}
    s_i  = & \sum_{k=1}^K  \probclassifiershort{} \cdot \log \probclassifiershort{}
    \\ 
    & - \sum_{k=1}^K \probempk{i} \cdot \log \probclassifiershort{}  \nonumber 
    % \label{eq:alc}
\end{align}

% This method is only designed for settings where all examples  have at least one annotation in the dataset (i.e.\ $\mathcal{U} = \emptyset$).

\begin{figure*}[tb]
\begin{subfigure}[t]{.5\textwidth}
  \centering
  % first image
  \includegraphics[width=\linewidth]{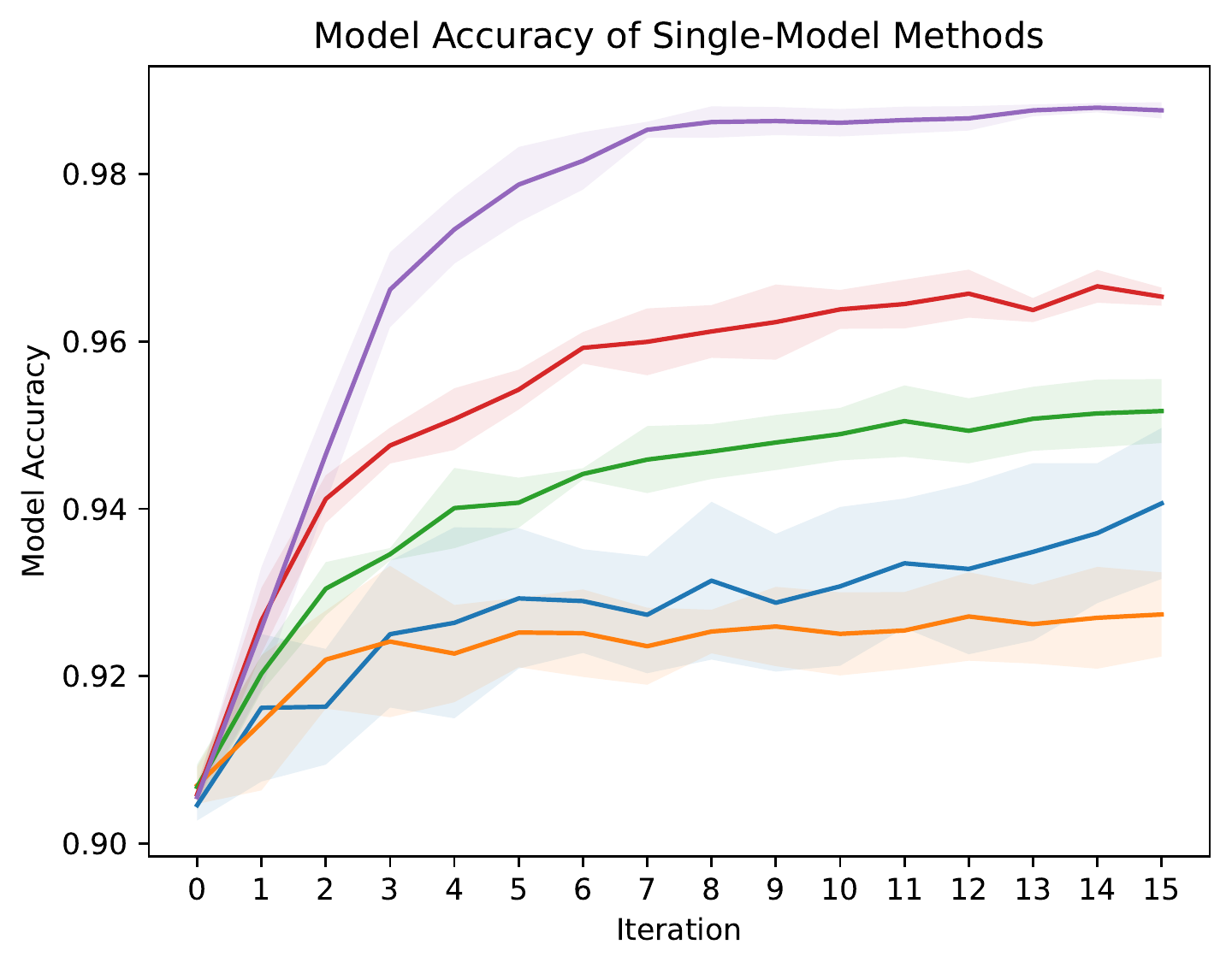} 
  % \caption{potential subcaption}
  \label{fig:wr_et_single}
    % \vspace*{-0.7em}
\end{subfigure}
\begin{subfigure}[t]{.5\textwidth}
  \centering
  % second image
  \includegraphics[width=\linewidth]{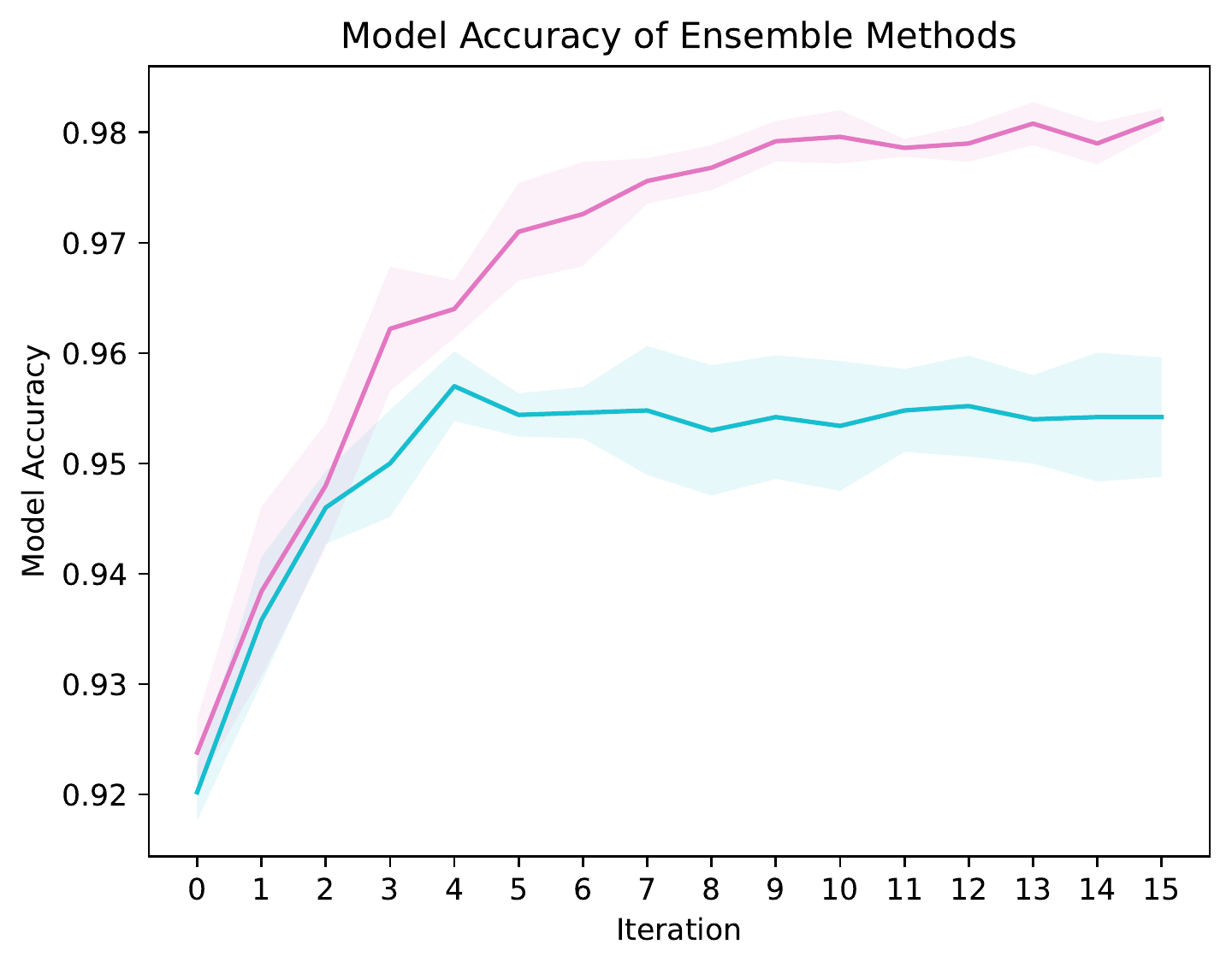} 
  % \caption{potential subcaption}
  \label{fig:wr_mlp_ensemble}
   %  \vspace*{-0.7em}
\end{subfigure}
\\[-1.4em]
\begin{subfigure}[t]{.5\textwidth}
  \centering
   \hspace*{1em}
  \includegraphics[width=\linewidth]{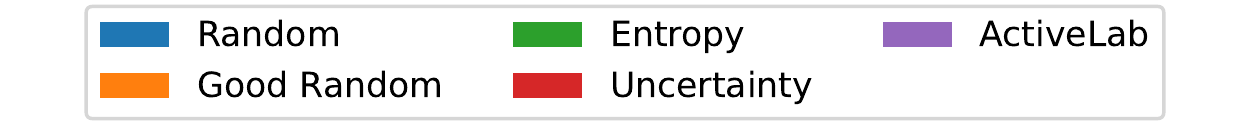} 
  \label{fig:wr_single_legend}
\end{subfigure} 
\begin{subfigure}[t]{.5\textwidth}
  \centering
  \hspace*{1em}
  \includegraphics[width=\linewidth]{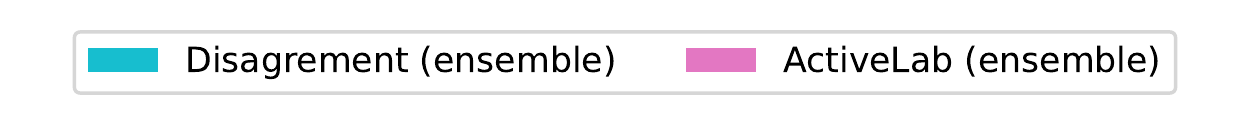} 
  \label{fig:wr_ensemble_legend}
  \vspace*{-6mm}
\end{subfigure} 
\vspace*{-5mm}
\caption{Evaluating active learning methods on the Wall Robot dataset to train an: ExtraTrees classifier (left) or ensemble of 3 models (right). 
Curves show test accuracy after each active learning iteration, averaged over 5 runs with the standard deviation in results shaded.
}
\label{fig:wr}
\end{figure*}

\textbf{Disagreement (Ensemble)} \cite{seung1992query}.  Like ActiveLab (Ensemble), \emph{disagreement} also employs an ensemble of multiple classifier models. This method measures the level of disagreement between different individual models' predictions. We employ a standard measure of disagreement for predicted class probabilities, where the score is defined as the total (soft) cross entropy between each model's predicted probabilities and the average estimate over all the models \cite{mccallum1998employing}.
\begin{gather}
    s_i = - \frac{1}{L} \sum_{\ell=1}^L \sum_{k=1}^K \probclassifierensembleshort{\ell} \cdot \probclassifiermeanshort \\ 
    \text{ where } \ \probclassifiermeanshort = \frac{1}{L} \sum_{\ell=1}^L \probclassifierensembleshort{\ell} \nonumber
\end{gather}
To produce predictions from our ensemble classifier after running this method, we simply average the predictions from the individual models.

\section{Experiments}
\label{sec:experiments}

% \subsection{Procedure}

In our experiments, each dataset is partitioned  into train, test, and unlabeled pools. 
We have high-quality (i.e.\ ground truth) labels for the test set, which facilitates accurate evaluation of trained classifiers. No such ground-truth labels are available for the training set. Instead, all examples in the training set have been labeled by one or more (potentially noisy) annotators, and we consider this to be the dataset $\mathcal{D}$ for training an initial classifier, collected prior to active learning.
At the outset, no labels are available for examples in the unlabeled pool. The train/test/unlabeled pools and the initial training annotations are identical across all runs/methods evaluated for the dataset. 
After training the model in Step \ref{step:train} of each round of active learning, we evaluate its test accuracy against ground truth labels (only used for evaluation purposes). 
To acquire labels in Step \ref{step:collect}, our experiments use a single new annotator to label the entire selected batch of data from a round of active learning.

\subsection{Datasets and Models}
\label{sec:dataset-models}

We evaluate active learning methods on datasets of different modalities and sizes, training various classification models for these datasets to ensure our methods are model agnostic.

\textbf{Wall Robot Navigation} \cite{wallrobot}.  This is a tabular dataset with 4 classes corresponding to directions a robot should navigate which are to be predicted from its sensor measurements. 
The initial train set for this dataset contains 500 examples, the unlabeled pool contains 1500 examples, and the test set used to measure the model accuracy contains 1000 examples. In each round of active learning between model training runs, we collect additional labels for the 100 examples with the lowest active learning scores from a single new annotator. We simulate imperfect annotators for this dataset. Some of these 100 examples may already have been previously labeled by other annotators and some may not have been labeled at all yet. 

We consider 3 types of classifier models: Extremely Randomized Trees (Extra Trees) \cite{geurts2006extremely}, which was the most accurate model from the \texttt{sklearn} package on this dataset,  fully-connected neural networks (MLP), K-Nearest Neighbors, and an ensemble composed of all 3. 
% We use an Extra Trees model to run all the experiments for the single-model methods (ie. the model accuracy presented is the model accuracy of this trained Extra Trees models on the test set). For the ensemble methods, we used three classifier models: Extra Trees, a Multilayer Perceptron and K-Nearest Neighbors. The predicted label from the ensemble model is obtained by taking the label that has the highest average predicted probability from the three ensemble models. That ensemble predicted label is the label that is used to compute the model accuracy for the ensemble methods.

\begin{figure*}[tb]
\begin{subfigure}[t]{.5\textwidth}
  \centering
  % first image
  \includegraphics[width=\linewidth]{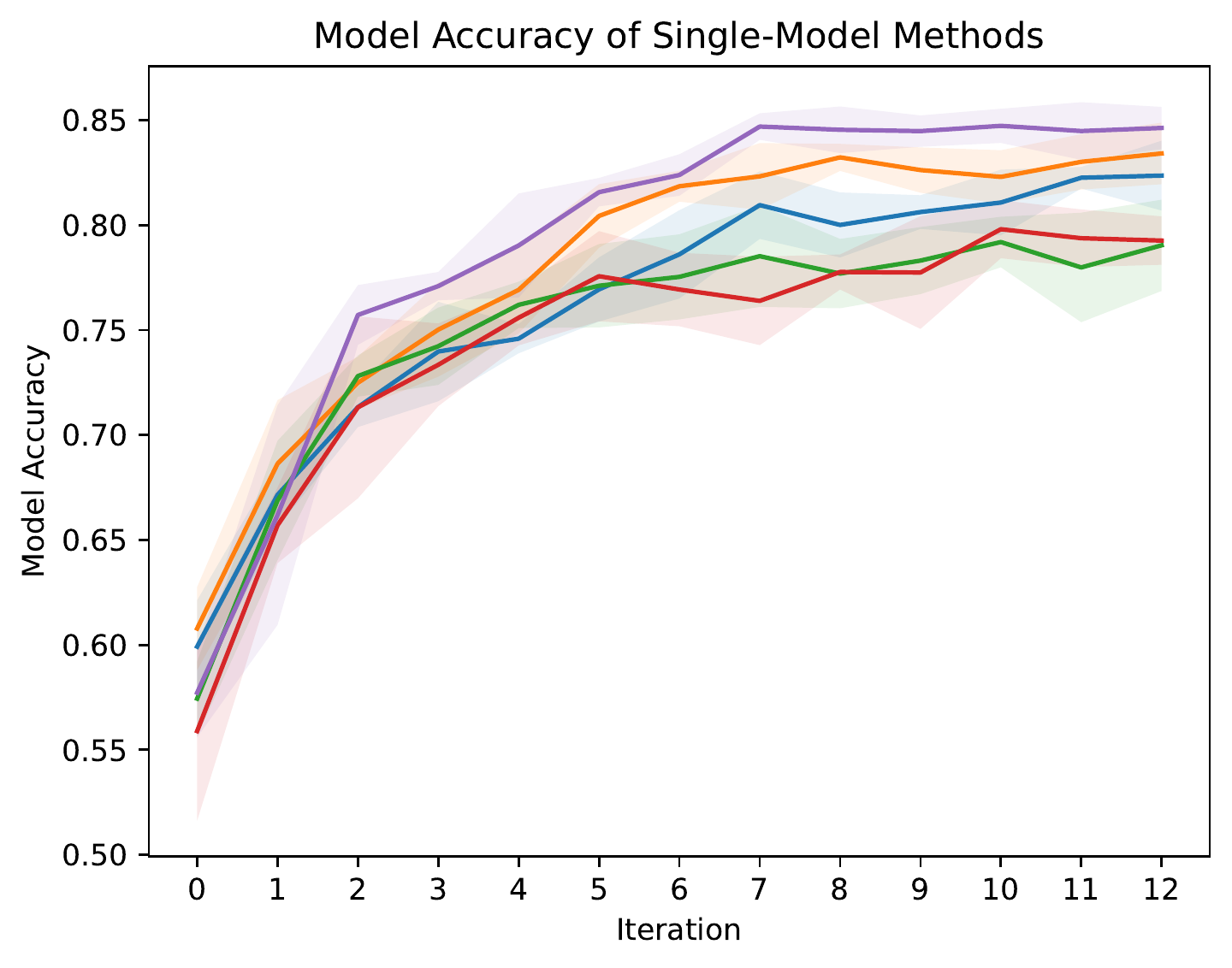} 
  % \caption{potential subcaption}
  \label{fig:cifar_single}
    % \vspace*{-0.7em}
\end{subfigure}
\begin{subfigure}[t]{.5\textwidth}
  \centering
  % second image
  \includegraphics[width=\linewidth]{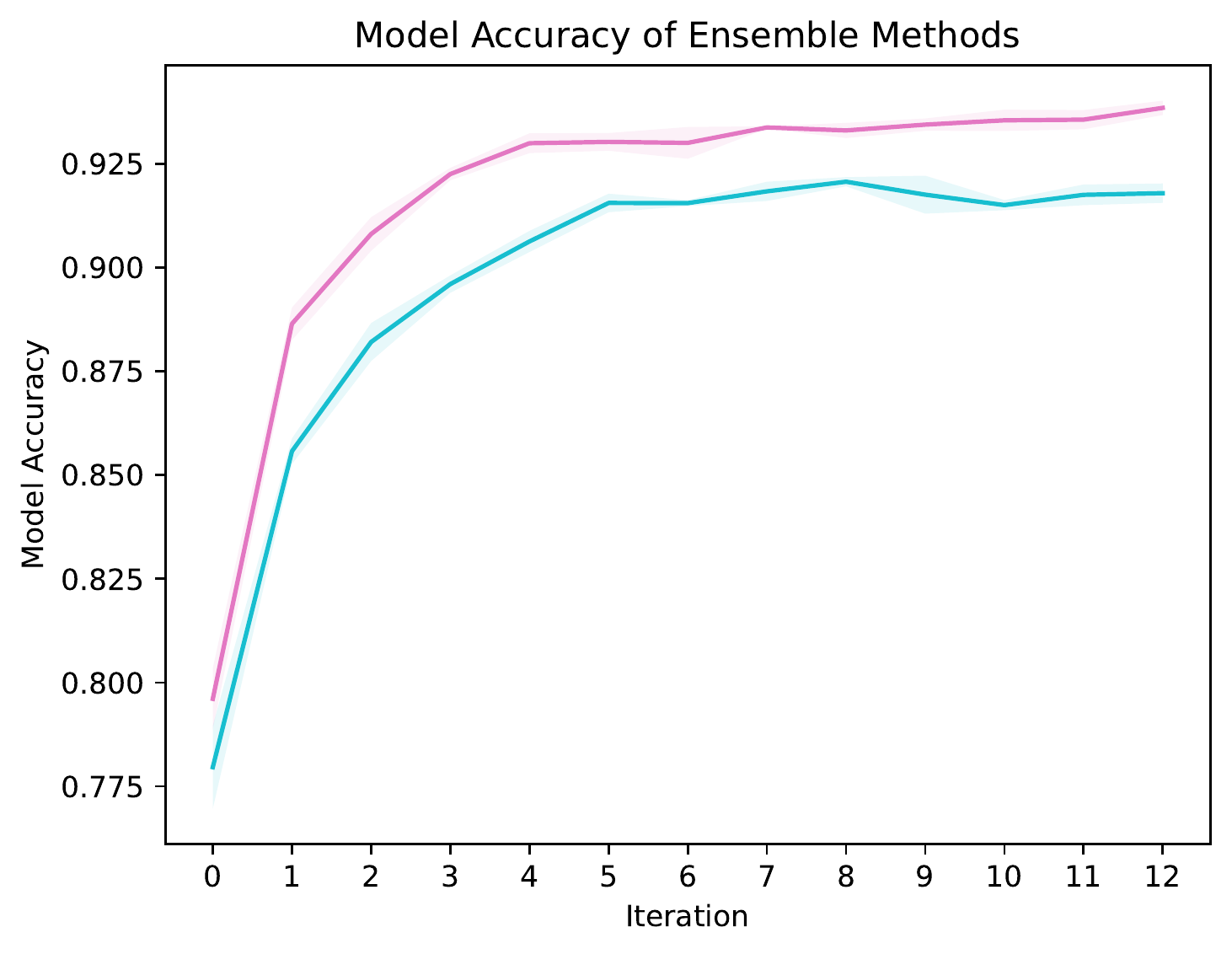} 
  % \caption{potential subcaption}
  \label{fig:cifar_ensemble}
    % \vspace*{-0.7em}
\end{subfigure}
\\[-1.4em]
\begin{subfigure}[t]{.5\textwidth}
  \centering
 \hspace*{1em} \includegraphics[width=\linewidth]{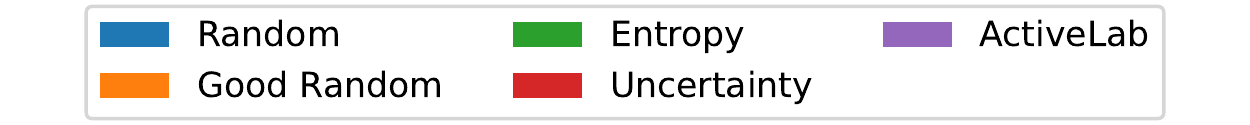} 
  \label{fig:cifar_single_legend}
\end{subfigure} 
\begin{subfigure}[t]{.5\textwidth}
  \centering
 \hspace*{1em} \includegraphics[width=\linewidth]{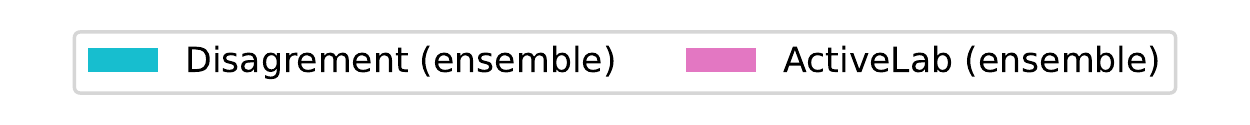} 
  \label{fig:cifar_ensemble_legend}
  \vspace*{-6mm}
\end{subfigure} 
\vspace*{-5mm}
\caption{Evaluating active learning methods on CIFAR-10H to train a: ResNet-18 classifier (left) or ensemble of ResNet-18/34/50 models. Curves show the test accuracy after each  iteration of active learning, averaged over 5 runs with the standard deviation in results shaded.}
\label{fig:cifar}
\end{figure*} 

\textbf{CIFAR-10H} \cite{cifar10h}.  This image classification dataset offers many annotated labels for each image in the CIFAR-10 test set, provided by different human annotators. Our experiment uses a subset of 1000 images as the initial training set, 4000 images in the unlabeled pool, and 5000 images in the test set. Our high-quality test set labels to measure model accuracy are those from the original CIFAR-10 dataset \cite{cifar10}, as \citet{northcutt2021labelerrors} found the CIFAR-10 labels contain few errors. In each round of active learning, we collect additional labels from one new human annotator for the 500 images with the lowest scores $s_i$. 

We use an Imagenet-pretrained ResNet-18 classifier for single-model active learning. For  ensemble-model active learning, our ensemble consists of three classifiers: ResNet-18, ResNet-34 and ResNet-50 \cite{he2016deep}.

\textbf{Wall Robot Complete}.  Similar to the Wall Robot Navigation tabular dataset, a key difference is that \emph{Wall Robot Complete} has 2000 labeled examples in the initial training set, 1000 examples in the test set, and there is no unlabeled pool. 
As for \emph{Wall Robot Navigation}, we collect additional labels for the 100 examples with the lowest active learning scores in each active learning round. Since all the examples already start out with some labels, this is a re-labeling (i.e.\ label cleaning) task, where we aim to obtain accurate   consensus labels by having multiple annotators review the examples where this is necessary \cite{bernhardt2022active}.

\section{Results}

Our main evaluation criterion is the test accuracy of classifier trained in each round of active learning.
% Although all active learning methods start with the same dataset at round 0, however there might be some discrepancy of the model accuracy at round 0 due to the non-deterministic nature of the classifier models.
Each experiment (sequential active learning run) is repeated 5 times and we report the average model accuracy across the trials.
% Here we present results of the benchmarks. To evaluate how well each method is performing, we measure the model accuracy of the classifier model on a held-out test dataset at each round of active learning. 

% \subsection{CIFAR-10H Results}

% \subsection{Wall Robot Complete Results}

\begin{figure*}
\begin{subfigure}[t]{.5\textwidth}
  \centering
  % first image
  \includegraphics[width=\linewidth]{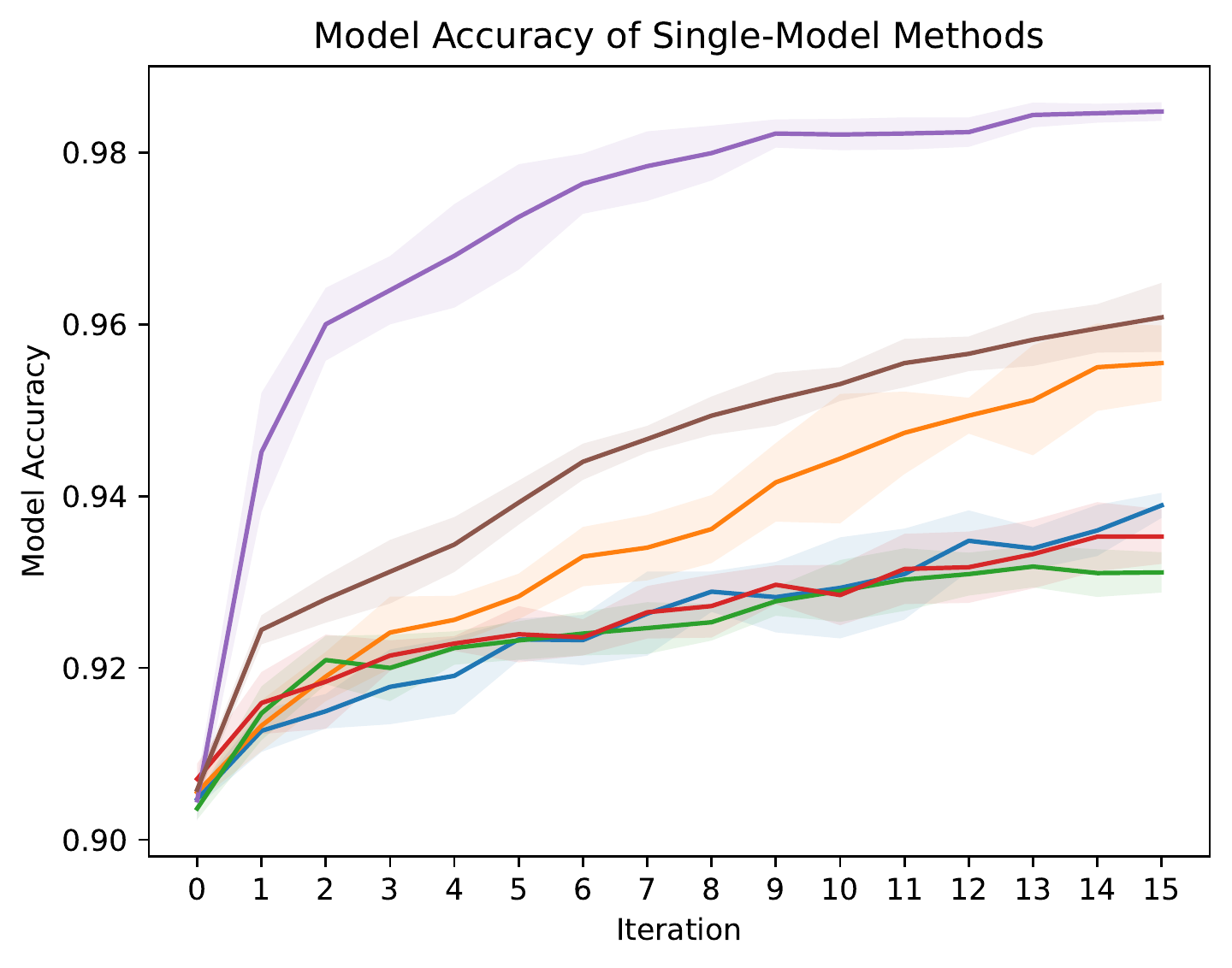} 
  % \caption{potential subcaption}
  \label{fig:wrc_et_single}
    % \vspace*{-0.7em}
\end{subfigure}
\begin{subfigure}[t]{.5\textwidth}
  \centering
  % second image
  \includegraphics[width=\linewidth]{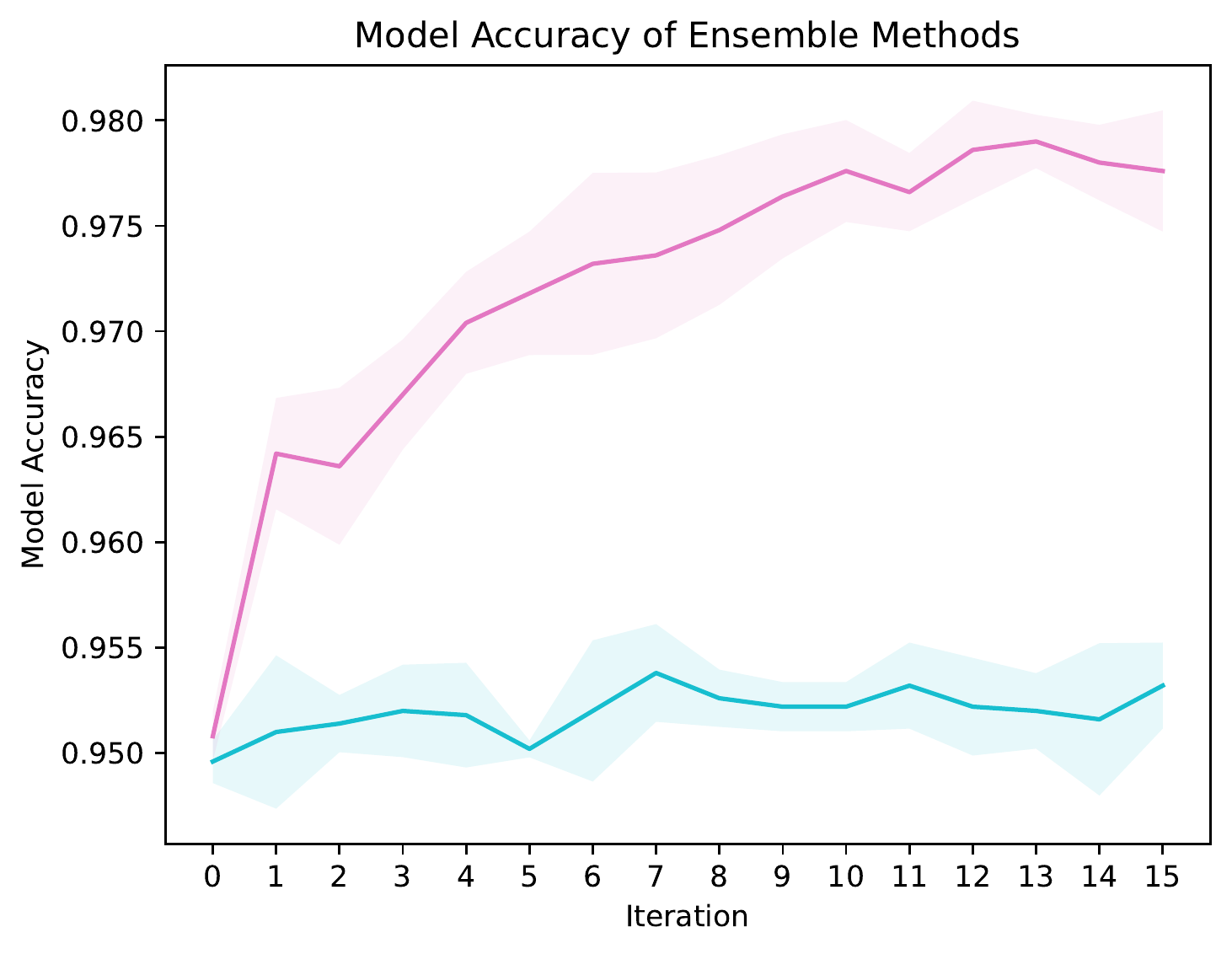} 
  % \caption{potential subcaption}
  \label{fig:wrc_mlp_ensemble}
    % \vspace*{-0.7em}
\end{subfigure}
\\[-1.4em]
\begin{subfigure}[t]{.5\textwidth}
  \centering
\hspace*{0.5em}  \includegraphics[width=\linewidth]{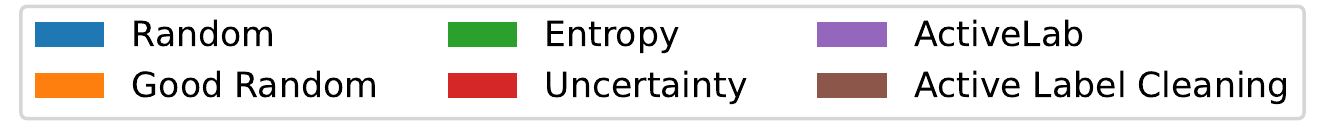}
  \label{fig:wrc_single_legend}
\end{subfigure}
\begin{subfigure}[t]{.5\textwidth}
  \centering
\hspace*{1em}  \includegraphics[width=\linewidth]{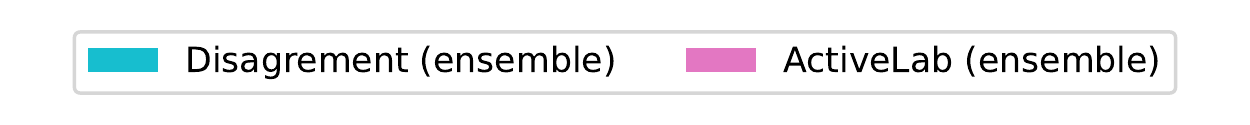} 
  \label{fig:wrc_ensemble_legend}
  \vspace*{-5mm}
\end{subfigure} 
\vspace*{-5mm}
\caption{
Evaluating active learning methods on the Wall Robot Complete dataset to train an: ExtraTrees classifier (left) or ensemble of 3 models (right). 
Curves show test accuracy after each iteration of re-labeling, averaged over 5 runs with the standard deviation shaded.}
\label{fig:wrc}
\end{figure*}

Figures \ref{fig:wr}, \ref{fig:cifar} and \ref{fig:wr_extra}  illustrate that ActiveLab significantly outperforms the other active learning methods in both the single-model and ensemble setting. These findings demonstrate that ActiveLab effectively selects examples to label and re-label in data of various modalities modeled with different types of classifiers.
Unsurprisingly, active learning with ensemble models can produce higher accuracy than achieved with single models.
Although note that single model accuracy when collecting data with ActiveLab can attain comparable performance to the ensemble models, especially for strong single models like in Figure \ref{fig:wr}
% When comparing the baseline methods, the ensemble methods perform better compared to the single-model methods on average (ie. the baseline disagreement method performs better than most single-model methods). Intuitively, this is as expected because we are able to extract more information from the multiple classifier models to determine the active learning quality score for each example. 
% The only exception to this is the CIFAR-10H experiments where both ensemble methods perform significantly better than the single-model methods. This can likely be attributed to the fact that in the single-model methods we used a pretrained ResNet-18 classifier, whereas in the ensemble methods we added pretrained ResNet-34 and ResNet-50 classifiers, both of which generally performs better than the smaller ResNet-18 classifier, hence resulting in a much higher recorded model accuracy. This also illustrates the importance of how a better classifier can impact the active learning quality score estimates. 

Figure \ref{fig:wrc} shows that ActiveLab is also the best method for active label cleaning (re-labeling an already labeled dataset). It even outperforms the method \citet{bernhardt2022active} designed specifically for this setting. Unlike \citet{bernhardt2022active}, ActiveLab estimates account for the number of annotations each example has and the quality of the annotators behind them. Existing active learning methods do not appear well-suited for such label cleaning tasks.
% is conducted on a dataset in which all examples already have labels. The results in this experiment show that traditional active learning methods, which are mostly built for collecting label from new examples, are not very well-suited for correcting incorrect labels in datasets (as seen by most methods having a similar small amount of model improvement). The Active Label Cleaning method, which focus is to collect more labels for already labeled examples to reduce label noise, performs better than the other baseline methods. However, as can be seen in the figure, the ActiveLab method further outperforms the Active Label Cleaning method. The ActiveLab method also takes into consideration the number of labels each examples has and the quality of the annotators that label each example, which the Active Label Cleaning method does not, hence leading to a better estimate of the trustworthiness of each example.

\subsection{Labeling New Examples vs Re-labeling}
\label{sec:single_vs_multi}

Traditional active learning only considers collecting at most one label per example and focuses entirely on the unlabeled pool rather than considering the option to re-label. If we have a huge unlabeled pool and a limited labeling budget, is there any utility in re-labeling? Our previous results clearly demonstrate the value of smart re-labeling when the size of $\mathcal{U}$ and labeling budgets are suitably matched. But with near-perfect annotators and an near-infinite unlabeled pool, re-labeling might not seem like a good idea \cite{lin2014re}. Thus we empirically investigate the question: At what degree of annotation-noise is there value in re-labeling when the size of $\mathcal{U}$ greatly exceeds our labeling budget?

We consider two settings: one where we only label \emph{new} examples in each active learning round (\emph{single label} case), and another where we can re-label examples if ActiveLab chooses to do so (\emph{multiannotator label} case). We run these approaches on a few variants of the Wall Robot Navigation dataset where we simulate annotators with different label noise rates. A higher noise rate annotator produces  labels which are often wrong, while an annotator with noise rate 0 always selects labels that are correct. Similar to our previous Wall Robot benchmark, we conduct this experiment  with an initial train set of 500 labeled examples, an unlabeled pool of 1500 examples, and test set of 1000 well-labeled examples. We label batches of 100 examples in each active learning round. 
Both \emph{single label} and \emph{multiannotator label} experiments start with the same labeled subset $\mathcal{D}$ (and always have the same annotator noise rates). In the \emph{single label} experiment, active learning is done using the traditional entropy score only considering examples in $\mathcal{U}$. In the \emph{multiannotator label} experiment,  active learning is done via ActiveLab, which often selects a mixture of examples from $\mathcal{D}$ and $\mathcal{U}$ to collect an additional label for. 

% Active learning is commonly done in the case where each example has only one label, and in each active learning round, new examples (that were previously unlabeled) are labeled. However, this paper presents the idea of re-labeling examples that have already been labeled with the aim of obtaining a more accuracy consensus label, hence correcting any label errors that might have previously been in the training data. 

% Here, we explore that with a limited labeling budget, whether labeling new examples or re-labeling existing examples (to potentially get a better consensus label) is more effective in improving model accuracy. The case where we label new examples each round will be referred to as the \emph{single label} case, as generally we will only have one label per examples for that scenario; while the case where we can re-label examples will be referred to as the \emph{multiannotator label} case.

\begin{figure}[H]
  \vspace*{5mm}
\begin{subfigure}[t]{.5\textwidth}
  \centering
  % first image
  \includegraphics[width=\linewidth]{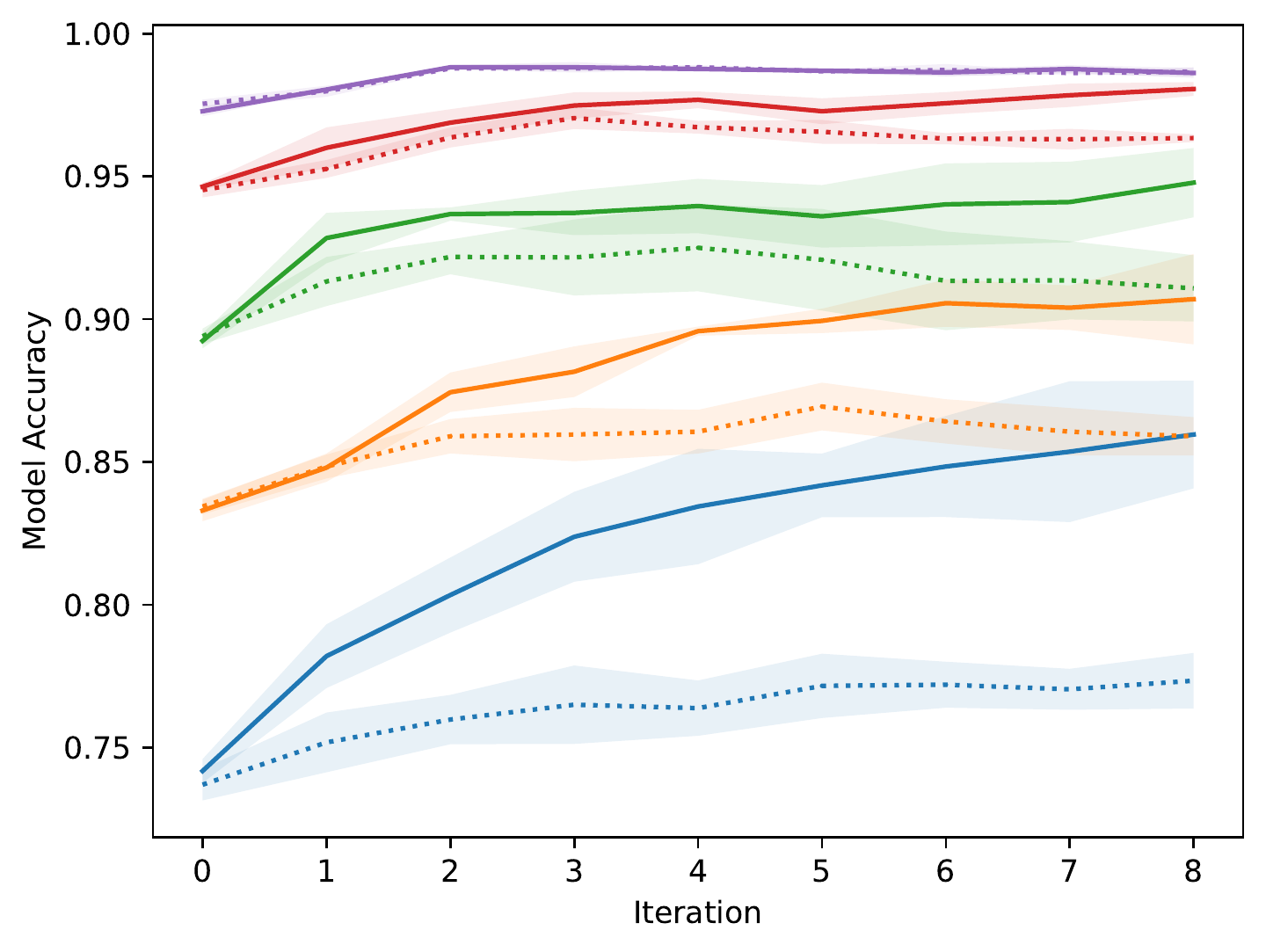} 
  % \caption{potential subcaption}
  \label{fig:single_vs_multi_acc}
\end{subfigure}
\\[-1em]
\begin{subfigure}[t]{.5\textwidth}
  \centering
  % second image
  \includegraphics[width=\linewidth]{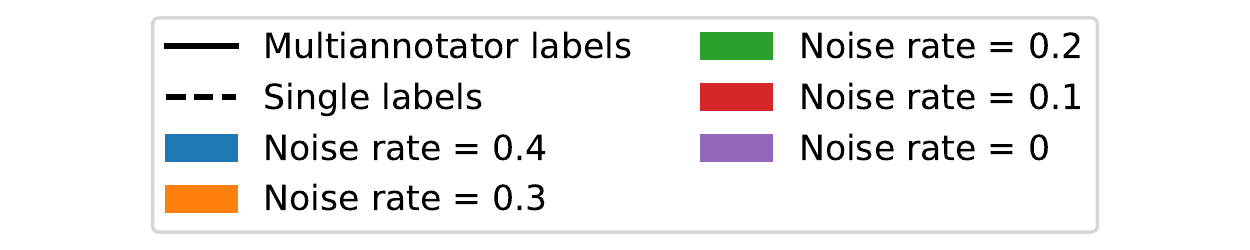} 
  % \caption{potential subcaption}
  \label{fig:single_vs_multi_legend}
\vspace*{-5mm}
\end{subfigure}
% \vspace*{-5mm}
\caption{Comparing active learning methods that exclusively label new examples (\emph{single labels}) vs.\ can also re-label examples instead (\emph{multiannotator labels}), when annotators have different noise rates. Shown is the test accuracy of an ExtraTrees classifier trained on a certain number of total labels (corresponding to each  iteration of active learning) for the Wall Robot Dataset. Curves are the average over 5 runs, and the standard deviation in results is shaded.}
\label{fig:single_vs_multi}
\end{figure}

Figure \ref{fig:single_vs_multi} reveals that across all annotator noise levels, the model accuracy for the \emph{mulitannotator labels} case is equal or better than for \emph{single labels}. As expected, the difference in model accuracy between \emph{single labels} and \emph{mulitannotator labels} is larger when annotators are more noisy. This suggests it is rarely a bad idea to allow re-labeling if you have a method to do it adaptively like ActiveLab. It appears vital to re-label in settings with over 20\% label noise. Our findings run contrary to the study of \citet{lin2014re}, who acknowledged they were missing an effective active learning method with re-labeling at the time of their study.

% \iffalse
\section{Discussion}

% As the technology for supervised learning continues rapidly advancing, these improvements in architectures and training procedures directly benefit model-agnostic active learning methods, which 

Relying on model predictions to infer what data is most informative, model-agnostic active learning methods will improve automatically as supervised learning architectures and training procedures continue to advance. More sophisticated active learning methods designed for specific models or training procedures will not enjoy these benefits and may become irrelevant if incompatible with tomorrow's  state-of-the-art models. 
Unlike traditional model-agnostic active learning that solely relies on model predictions to determine which examples to label next, ActiveLab considers re-labeling examples $x_i \in \mathcal{D}$ and estimates the value of this based on additional information like the: number of available annotations for $x_i$, disagreement amongst these annotations, and relative trustworthiness of the trained model vs.\ the annotators. Re-labeling facilitates more robust model training when data annotators are imperfect. Future work might seek to achieve further robustness by filtering bad annotators (e.g.\ based on their weights $w_j$) and data/labels from the training set $\mathcal{D}$ of each active learning round.

\clearpage \newpage

% \let\oldthebibliography\thebibliography
% \let\endoldthebibliography\endthebibliography
% \renewenvironment{thebibliography}[1]{
%   \begin{oldthebibliography}{#1}
%     \setlength{\itemsep}{1.5em}
%     \setlength{\parskip}{0em}
% }
% {
%   \end{oldthebibliography}
% }

\bibliographystyle{icml2023}
\bibliography{activelearning}
\balance{}

%%%%%%%%%%%%%%%%%%%%%%%%%%%%%%%%%%%%%%%%%%%%%%%%%%%%%%%%%%%%%%%%%%%%%%%%%%%%%%%
%%%%%%%%%%%%%%%%%%%%%%%%%%%%%%%%%%%%%%%%%%%%%%%%%%%%%%%%%%%%%%%%%%%%%%%%%%%%%%%
% APPENDIX
%%%%%%%%%%%%%%%%%%%%%%%%%%%%%%%%%%%%%%%%%%%%%%%%%%%%%%%%%%%%%%%%%%%%%%%%%%%%%%%
%%%%%%%%%%%%%%%%%%%%%%%%%%%%%%%%%%%%%%%%%%%%%%%%%%%%%%%%%%%%%%%%%%%%%%%%%%%%%%%
\clearpage \newpage
\beginsupplement
\onecolumn
\appendix

\def\toptitlebar{\hrule height1pt \vskip .16in} 
\def\bottomtitlebar{\vskip .22in \hrule height1pt \vskip .3in} 

\thispagestyle{plain}

\setcounter{page}{1}
\pagenumbering{arabic}
\setlength{\footskip}{20pt}  
% \vspace*{-3.5mm}
\begin{center}
\toptitlebar
{\Large \bf \appendixtitle 
}
\bottomtitlebar
\end{center}

\FloatBarrier

\section{Details of the Wall Robot Navigation Dataset}

The original Wall-Following Robot Navigation dataset only has one label for each example. 
We adapt this dataset for our multi-annotator benchmark by simulating many different annotators to provide labels for requested examples. 
To simulate human annotators that make imperfect decisions  (i.e.\ occasional labeling errors), we take the original set of labels from the Wall Robot dataset as ground truth labels. For each annotator, we add some random noise to their labels (noise rate = 0.15 for \emph{Wall Robot Navigation} and noise rate = 0.2 for \emph{Wall Robot Complete}), representing mislabeled examples. The randomly selected noisy annotations have an incorrect class (flipped probabilistically) that does not match the ground truth label. Using this method, we obtained 30 sets of labels, representing 30 annotators. 

To setup the initial labeled and unlabeled pools $\mathcal{D}$ and $\mathcal{U}$, we completely dropped all the annotator labels for examples that begin unlabeled, while dropping a random fraction of the annotator labels for the examples in $\mathcal{D}$ that are labeled from the start, ensuring we keep at least one annotation for these examples. When collecting additional labels in each round of active learning, we simulate another annotator in the same fashion who labels the entire batch.

We also considered a second version of this benchmark with more heterogeneous annotators (including some very inaccurate outliers), and the results of the evaluation remained mostly the same as those presented here.

\section{Experiment Details}

In each round of active learning, we fit all models to $\mathcal{D}$ using 5-fold cross-validation. Additional details not mentioned here can be found in the code\footnote{\gitlink{}} for reproducing our experiments, as can raw the results of all active learning methods on all datasets.

For the experiments on the tabular Wall Robot dataset, we fit our classifier models using the \texttt{sklearn} package \cite{scikit-learn}. The models used were the:  ExtraTreesClassifier with default hyperparameters,  MLPClassifier with default hyperparameters except the max iteration set to 500 (to ensure convergence), and  KNeighborsClassifier with default hyperparameters. 

The image classifier models for our experiments on CIFAR-10H were fit using the AutoGluon AutoML package \cite{agtabular} in order to avoid having to manually tune models and their optimization. We used various ResNet models initialized with default Imagenet-pretrained weights and then fine-tuned them on our dataset $\mathcal{D}$ (in a cross-valdated manner).

\section{Additional Results for Wall Robot}

In addition to the Extra Trees model reported in Section \ref{sec:dataset-models}, we repeat our single-model active learning experiments on the Wall Robot dataset using a Multilayer Perceptron (feedforward neural network) classifier. These additional results demonstrate that ActiveLab reliably produces larger improvement in model accuracy than other active learning methods, regardless which type of classifier model is being trained. 
 
% Below we will present the results for that experiment.

\begin{figure}[H]
\begin{subfigure}{.7\textwidth}
  \centering
  % first image
  \includegraphics[width=\linewidth]{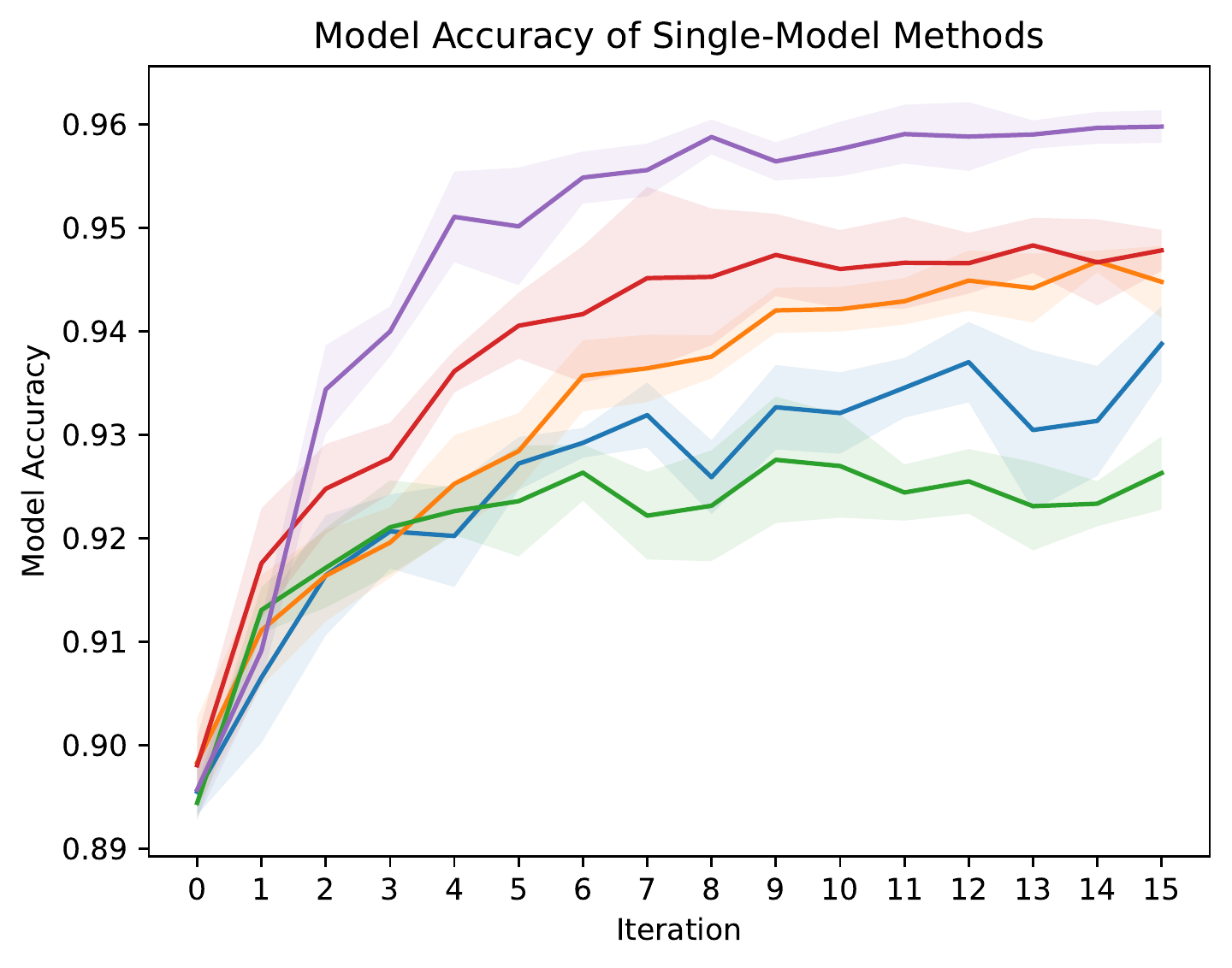} 
  % \caption{potential subcaption}
  \label{fig:wr_mlp_single_label}
\end{subfigure}
\begin{subfigure}{.3\textwidth}
  \centering
  % second image
  \includegraphics[width=\linewidth]{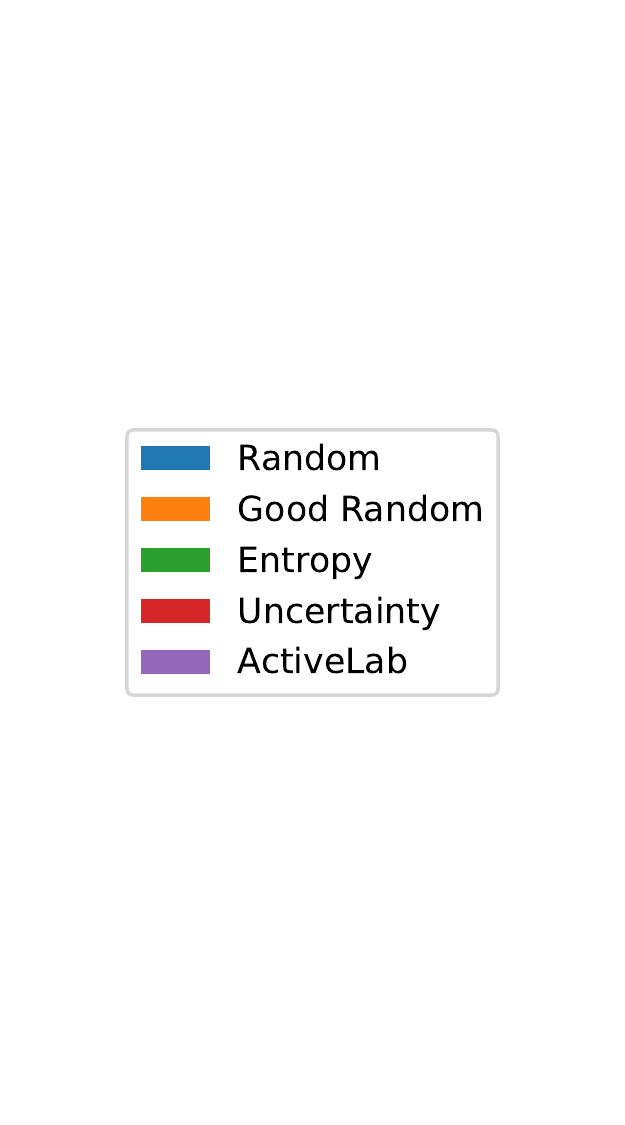} 
  % \caption{potential subcaption}
  \label{fig:wr_single_label_legend}
    \vspace*{-10mm}
\end{subfigure}
\vspace*{-10mm}

\caption{Evaluating active learning methods on the Wall Robot dataset to train a MLP classifier. 
Curves show the test accuracy after each active learning iteration, averaged over 5 runs with the standard deviation in results shaded.}
\label{fig:wr_extra}
\end{figure}

\begin{figure}[H]
\begin{subfigure}{.7\textwidth}
  \centering
  % first image
  \includegraphics[width=\linewidth]{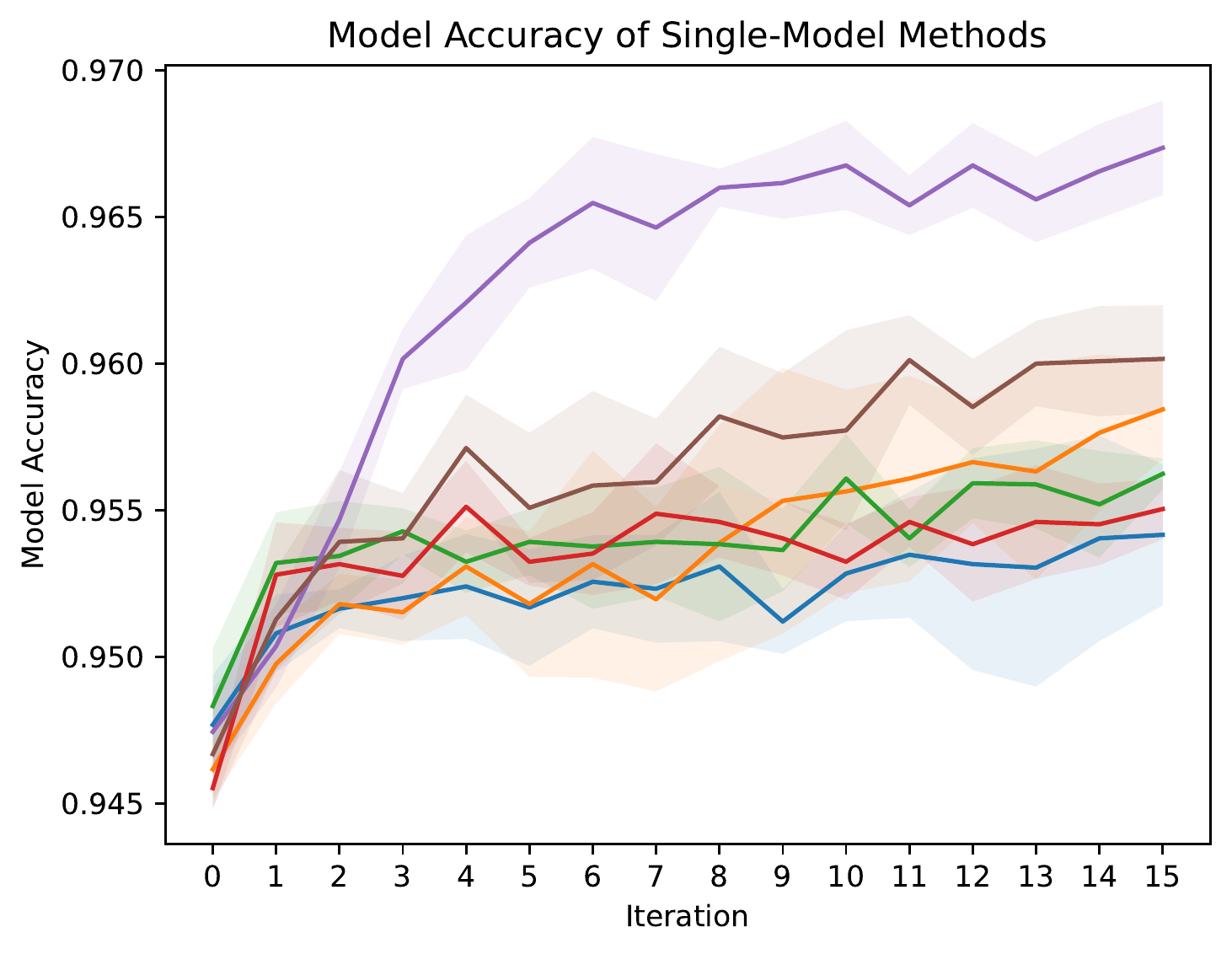} 
  % \caption{potential subcaption}
  \label{fig:wrc_mlp_single_label}
\end{subfigure}
\begin{subfigure}{.3\textwidth}
  \centering
  % second image
  \includegraphics[width=\linewidth]{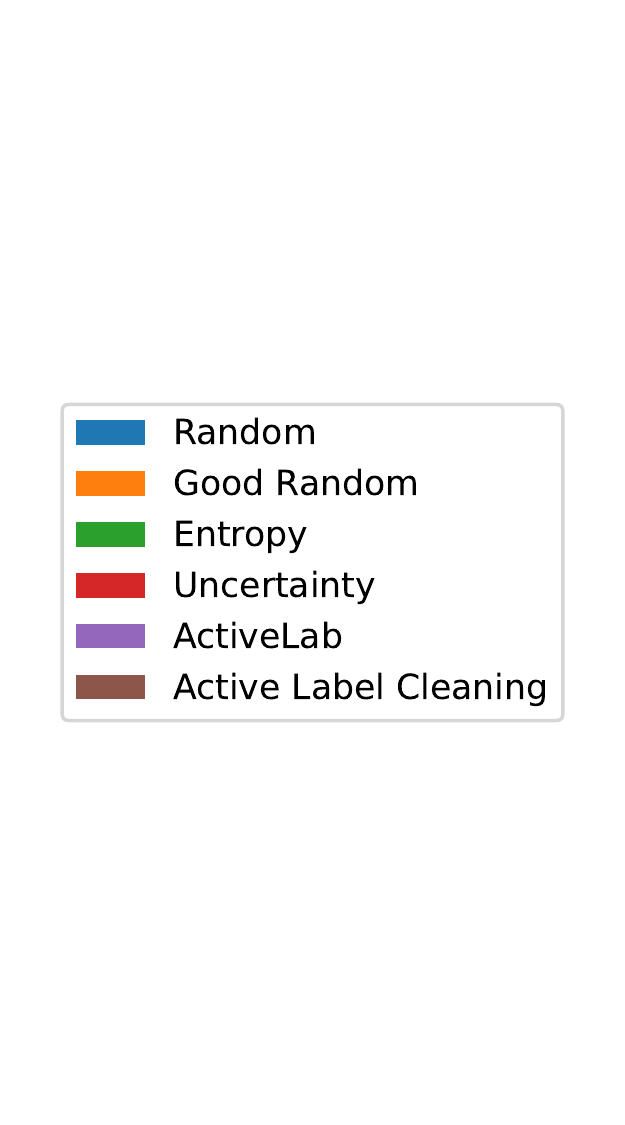} 
  % \caption{potential subcaption}
  \label{fig:wrc_single_label_legend}
  \vspace*{-10mm}
\end{subfigure}
\vspace*{-10mm}
\caption{Evaluating active learning methods on the Wall Robot Complete dataset to train a MLP classifier. 
Curves show the test accuracy after each iteration of re-labeling, averaged over 5 runs with the standard deviation shaded.}
\label{fig:wrc_extra}
\end{figure}

\clearpage
\section{Active Learning in Single-Label Settings}
\label{sec:singlelabel}

While ActiveLab is designed for scenarios where multiple annotators can label the same example, the method can also be applied for traditional active learning settings where we collect at most one label for each example. In this singly-labeled setting, we only score the $x_i \in \mathcal{U}$, as is common practice in pool-based active learning.
% and we would like to only label examples that have not been previously labeled, which is the common setup in traditional active learning tasks.

In such settings, we do not have data from multiple annotators to estimate the relative trustworthiness of the annotators and our model. Thus ActiveLab weights are undefined, but they are also not needed since we are only scoring unlabeled data without annotations in this setting. As a result, the natural ActiveLab score in such settings is:
\begin{equation}
    s_i =
    \frac{ \max_k \probclassifiershort{} + \frac{1}{K}}{2} \ \ \text{ for } x_i \in \mathcal{U}
\end{equation}
This is equivalent to simply relying on the confidence of the classifier, and thus equivalent to selecting examples via the aforementioned \emph{Uncertainty} baseline method, a classic technique for active learning \cite{cohn1996active, munro2021human}. 
% to obtain a consensus label from, and hence we also do not have a way to measure the reliability of the annotator or the model. Hence, in the single-label case, we will not use any annotator or model weights as we did in section \ref{sec:activelab}. In the single label case, we are only interested in the subset of unlabeled examples (because we are not re-labeling examples), and which to label first within that subset. Hence we show how the active learning quality score is computed for the unlabeled examples:

In this singly-labeled setting, we provide a benchmark of this active learning method against two alternatives:  randomly selecting examples to label, or using the entropy score. 
We use the same version of the Wall Robot Navigation dataset as our other benchmarks (with a noisy annotator). The initial train set contains 500 examples, and there are  1500 examples in the unlabeled pool. Each round, we select the 100 unlabeled examples with the lowest score to label and add them to the labeled subset. After training, model accuracy is similarly measured on a held-out test set of  1000 examples.

\begin{figure}[H]
\begin{subfigure}{.7\textwidth}
  \centering
  % first image
  \includegraphics[width=\linewidth]{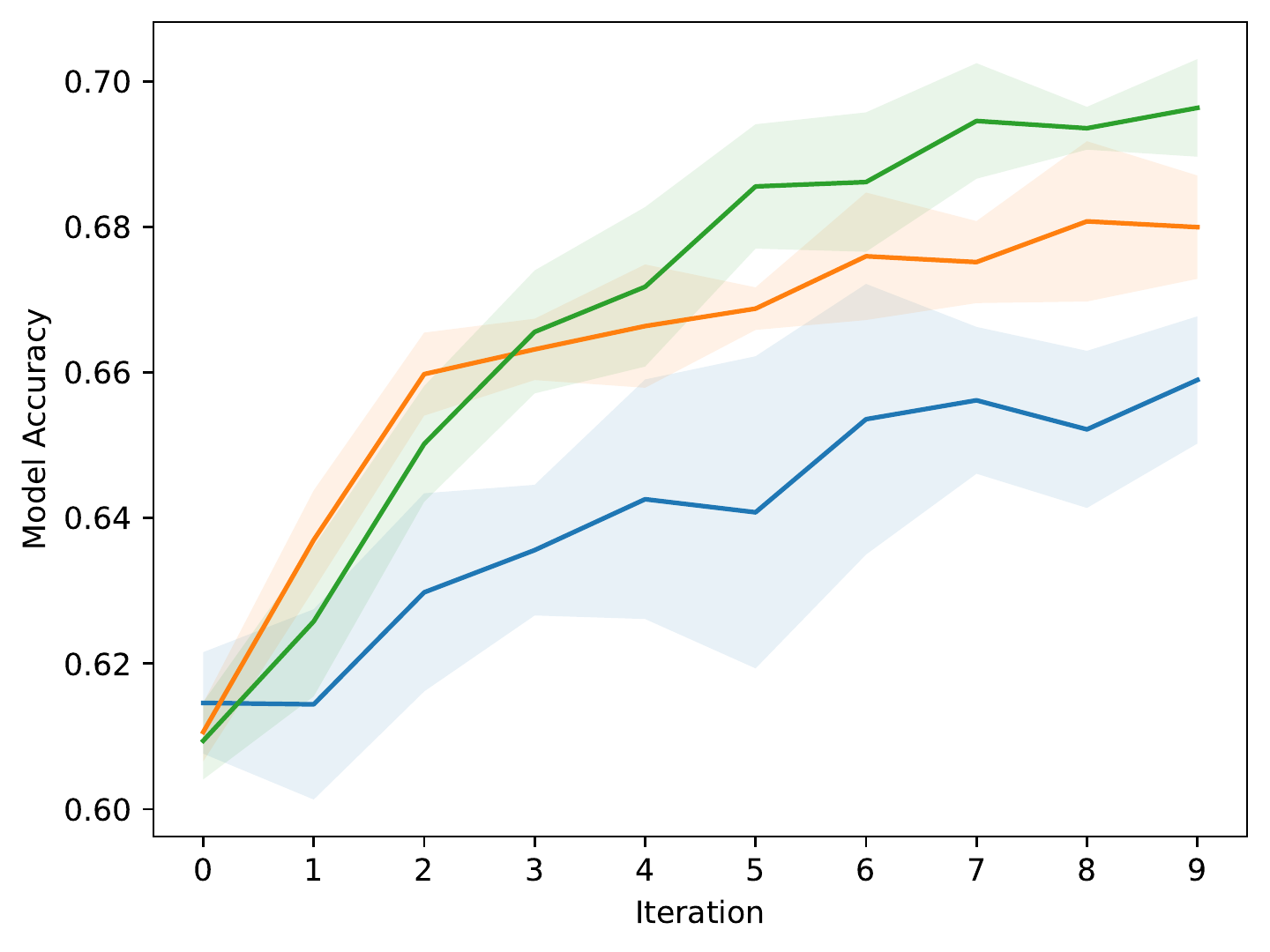} 
  % \caption{potential subcaption}
  \label{fig:single_label_acc}
\end{subfigure}
\begin{subfigure}{.3\textwidth}
  \centering
  % second image
  \includegraphics[width=\linewidth]{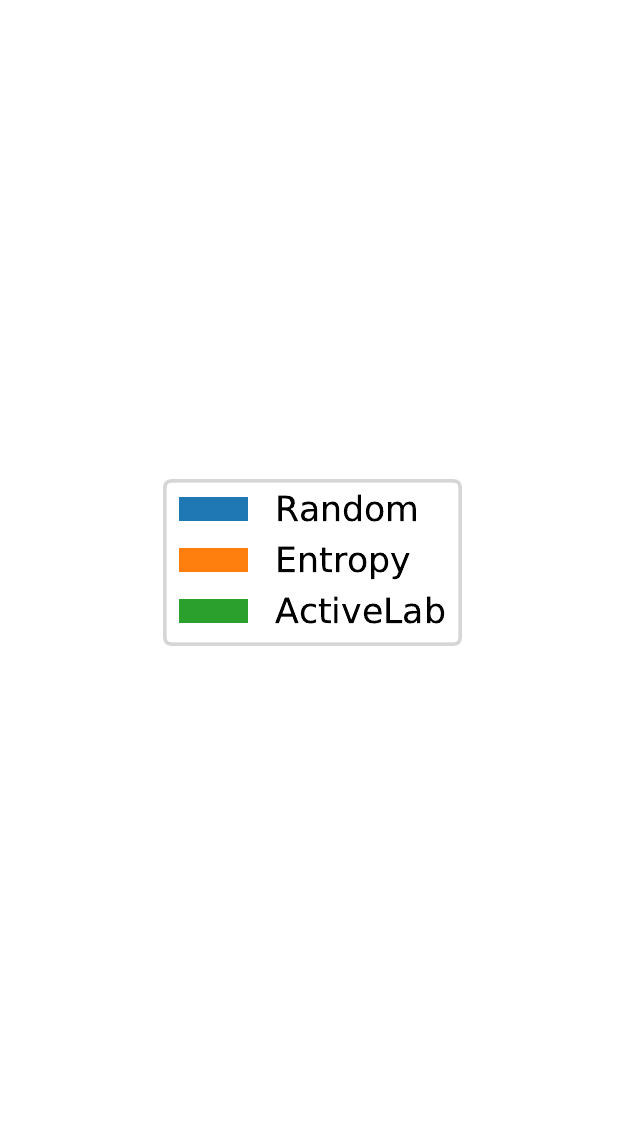} 
  % \caption{potential subcaption}
  \label{fig:single_label_legend}
\end{subfigure}
\vspace*{-10mm}
\caption{Evaluating active learning methods in the traditional singly-labeled setting on the Wall Robot dataset to train an ExtraTrees classifier.
Curves show the test accuracy after each active learning iteration, averaged over 5 runs with the standard deviation  shaded.}
\vspace*{-10mm}
\label{fig:single_label}
\end{figure}

Figure \ref{fig:single_label} shows that ActiveLab exhibits comparable performance to the entropy baseline method in the singly-labeled setting. Both methods significantly  outperform random selection of examples, even in this setting with noisy labels.

\end{document}